\documentclass[nohyperref]{article}

\usepackage{microtype}
\usepackage{graphicx}
\usepackage{subfigure}
\usepackage{booktabs} 

\usepackage{hyperref}
\usepackage{makecell}

\usepackage[accepted]{icml2022}

\usepackage{amsmath}
\usepackage{amssymb}
\usepackage{mathtools}
\usepackage{amsthm}

\usepackage[capitalize,noabbrev]{cleveref}

\theoremstyle{plain}

\theoremstyle{definition}

\theoremstyle{remark}

\usepackage[textsize=tiny]{todonotes}

\usepackage{amsfonts}
\usepackage{xr}

\usepackage{algorithm}
\usepackage{algorithmic}

\newcommand{\x}{\mathbf{x}}
\newcommand{\y}{\mathbf{y}}
\newcommand{\z}{\mathbf{z}}
\newcommand{\W}{\mathbf{W}}
\newcommand{\A}{\mathbf{A}}
\newcommand{\B}{\mathbf{B}}
\renewcommand{\b}{\mathbf{b}}
\renewcommand{\a}{\mathbf{a}}
\renewcommand{\A}{\mathbf{A}}
\renewcommand{\S}{\mathbf{S}}
\newcommand{\s}{\mathbf{s}}
\newcommand{\E}{\mathbf{E}}
\newcommand{\e}{\mathbf{e}}
\newcommand{\G}{\mathbf{G}}
\newcommand{\g}{\mathbf{g}}
\newcommand{\F}{\mathbf{F}}
\newcommand{\V}{\mathbf{V}}
\renewcommand{\H}{\mathbf{H}}
\newcommand{\I}{\mathbf{I}}
\newcommand{\U}{\mathbf{U}}
\renewcommand{\u}{\mathbf{u}}
\renewcommand{\v}{\mathbf{v}}
\newcommand{\w}{\mathbf{w}}

\newcommand{\0}{\mathbf{0}}

\newcommand{\EE}{\mathbb{E}}
\newcommand{\N}{\mathcal{N}}
\newcommand{\mat}{\mathrm{mat}}

\newcommand{\La}{\mathbf{\Lambda}}
\newcommand{\priorprec}{\La_\text{prior}}

\newcommand{\Sig}{\mathbf{\Sigma}}

\renewcommand{\vec}[1]{\mathrm{vec}\left( #1 \right)}
\newcommand{\nocontentsline}[3]{}
\newcommand{\tocless}[2]{\bgroup\let\addcontentsline=\nocontentsline#1{#2}\egroup}

\begin{document}

\icmltitlerunning{Gradient Descent on Neurons and its Link to Approximate Second-Order Optimization}	
\twocolumn[
\icmltitle{Gradient Descent on Neurons\\and its Link to Approximate Second-Order Optimization}

\begin{icmlauthorlist}
	\icmlauthor{Frederik Benzing}{yyy}
\end{icmlauthorlist}

\icmlaffiliation{yyy}{Department of Computer Science, ETH Zurich, Zurich, Switzerland}
\icmlcorrespondingauthor{Frederik Benzing}{benzingf@inf.ethz.ch}

\icmlkeywords{Machine Learning, ICML}

\vskip 0.3in
]



\printAffiliationsAndNotice{}  
%
%
\begin{abstract}%
	Second-order optimizers are thought to hold the potential to speed up neural network training, but 
	due to the enormous size of the curvature matrix, they typically require approximations to be computationally tractable. The most successful family of approximations are Kronecker-Factored, block-diagonal curvature estimates (KFAC). 
	Here, we combine tools from prior work to evaluate exact second-order updates with careful ablations to establish a surprising result: Due to its approximations, KFAC is not closely related to second-order updates, and in particular, it significantly outperforms true second-order updates.
	This challenges widely held believes and immediately raises the question why KFAC performs so well. Towards answering this question we present evidence strongly suggesting that KFAC approximates a first-order algorithm, which performs gradient descent on neurons rather than weights. Finally, we show that this optimizer often improves over KFAC in terms of computational cost and data-efficiency.   
\end{abstract}

\tocless{\section{Introduction}}

Second-order information of neural networks is of fundamental theoretical interest and has important applications in a number of contexts like optimization, Bayesian machine learning, meta-learning, sparsification and continual learning \cite{ lecun1990optimal,   hochreiter1997flat, mackay1992practical, bengio2000gradient, martens2010deep, grant2018recasting, sutskever2013importance, dauphin2014identifying,  blundell2015weight, kirkpatrick2017overcoming, graves2011practical}. However, due to the enormous parameter count of modern neural networks working with the full curvature matrix is infeasible and this has inspired many approximations. 
Understanding how accurate known approximations are and developing better ones is an important topic at the intersection of theory and practice. 

A family of approximations that has been particularly successful are Kronecker-factored, block diagonal approximations of the curvature. 
Originally proposed in the context of optimization \cite{martens2015optimizing}, where they have lead to many further developments \cite{grosse2016kronecker, ba2016distributed, desjardins2015natural, botev2017practical, george2018fast, martens2018kronecker, osawa2019large, bernacchia2019exact, goldfarb2020practical}, they have also proven influential in various other contexts like Bayesian inference, meta learning and continual learning \cite{ritter2018online, ritter2018scalable,  dangel2020modular, zhang2018noisy, wu2017scalable, grant2018recasting}. 

Here, we describe a surprising discovery: Despite its motivation, the KFAC optimizer does not rely on second-order information; in particular it significantly outperforms exact second-order optimizers. We establish these claims through a series of careful ablations and control experiments and build on prior work, which shows that exact second-order updates can be computed efficiently and exactly, if the dataset is small or when the curvature matrix is subsampled \cite{ren2019efficient, agarwal2019efficient}.

Our finding that KFAC does not rely on second-order information immediately raises the question why it is nevertheless so effective. To answer this question, we present evidence that KFAC approximates a different, first-order optimizer, which performs gradient descent in neuron- rather than weight space. We also show that this optimizer itself often improves upon KFAC, both in terms of computational cost as well as progress per parameter update.

\textbf{Structure of the Paper.} In Section~\ref{sec:background} we provide background and define terminology. The remainder of the paper is split into two parts. In Section~\ref{sec:nat_kfac} we carefully establish that KFAC is not closely related to second-order information. In Section~\ref{sec:foof} we introduce gradient descent on neurons (``FOOF'') and present evidence that KFAC's performance relies on similarity to FOOF and that FOOF offers further performance improvements.\clearpage

\tocless{\section{Background and Efficient Subsampled Natural Gradients}
	\label{sec:background}}
The most straight-forward definition of a ``curvature matrix'' is the Hessian $\H$ of the loss with respect to the parameters. However, in most contexts (e.g.\ optimization or Laplace posteriors), it is necessary or desirable to work with a positive definite approximations of the Hessian, i.e.\ an approximation of the form $\H \approx \G\G^T$; examples for such approximations include the (Generalised) Gauss Newton matrix and the Fisher Information. For simplicity, we will now focus on the Fisher, but our methods straightforwardly apply to any case where the columns of $\G$ are Jacobians.
The Fisher $\F$ is defined as 
\begin{align}\label{eq:fisher}
	\F = \EE_{X\sim\mathcal{X}} \EE_{y\sim p(\cdot \mid X,\w)} \left[\g(X,y) \g(X,y)^T\right]
\end{align}
where $X\sim\mathcal{X}$ is a sample from the input distribution, $y\sim p(\cdot \mid X, \w)$ is a sample from the model's output distribution (rather than the label given by the dataset, see \citet{kunstner2019limitations} for a discussion of this difference). $\g(X,y)$ is the "gradient", i.e.\ the (columnised) derivative of the negative log-liklihood of $(X,y)$ with respect to the model parameters $\w\in\mathbb{R}^{n}$. 

The natural gradient method preconditions normal first-order updates $\v$ by the inverse Fisher. Concretely, we update parameters in the direction of $(\lambda \I +\F)^{-1}\v$. Here, $\lambda$ is a damping term and can be seen as establishing a trust region.
Natural Gradients were proposed by Amari and colleagues, see e.g.~\cite{amari1998natural} and were motivated from an information geometric perspective. The Fisher is equal to the Hessian of the negative log-likelihood under the model's output distribution and thereby closely related to the standard Hessian \cite{martens2014new, pascanu2013revisiting}, so that Natural Gradients are typically viewed as a second-order method \cite{martens2015optimizing}. 

\subsection{Subsampled, Exact Natural Gradients}
If the dataset is moderately small, or if the Fisher is subsampled, i.e.\ evaluated on a mini-batch, then natural gradients can be computed exactly and efficiently as shown by \citet{ren2019efficient} with ideas described independently by \citet{agarwal2019efficient}. The key insight is to apply the Woodburry matrix inversion lemma and to realise that many intermediate quantities do not need to be stored or computed explicitly. 
We propose some modest theoretical as well as practical improvements to these techniques, which are deferred to Appendix~\ref{sec:nat_details} along with implementation details.



We also show that with an additional trick \cite{thanks_arnaud, hoffman1991constrained}, one can sample efficiently and exactly from the Laplace posterior, see Appendix~\ref{sec:applications}.   

A more detailed summary of related work can be found in Appendix~\ref{sec:related}.
\newpage

\tocless{\subsection{Notation and Terminology}
	\label{sec:notation}}

We typically focus on one layer of a neural network. For simplicity of notation, we consider fully-connected layers, but results can easily be extended to architectures with parameter sharing, like CNNs or RNNs. 

We denote the layer's weight matrix by $\W\in\mathbb{R}^{n \times m}$ and its input-activations (after the previous' layer nonlinearity) by $\A\in\mathbb{R}^{m\times D}$, where $D$ is the number of datapoints. The layer's output activations (before the nonlinearity) are equal to $\B = \W\A \in\mathbb{R}^{n\times D}$  and we denote the partial derivates of the loss $L$ with respect to these outputs (usually computed by backpropagation) by $\E = \frac{\partial L}{\partial \B}$.  If the label is sampled from the model's output distribution, as is the case for the Fisher \eqref{eq:fisher}, we will use $\E_F$ rather than $\E$.

We use the term ``\textbf{datapoint}'' for a pair of input and label $(X,y)$. In the context of the Fisher information, the label will always be sampled from the model's output distribution, see also eq \eqref{eq:fisher}. Note that with this definition, the total number of datapoints is the product of the number of inputs and the number of labels. 

Following \citet{martens2015optimizing}, the Fisher will usually be approximated by sampling one label for each input. For some controls, we will distinguish whether one label is sampled or whether the full Fisher is computed, and we will refer to the former as \textbf{MC Fisher} and the latter as \textbf{Full Fisher}.  

As is common in the ML context, we will use the term ``\textbf{second-order method}'' for algorithms that use (approximate) second derivatives.  The term ``\textbf{first-order method}'' will refer to algorithms which only use first derivatives or quantities that are independent of the loss, i.e.\ ``zero-th'' order terms.


\tocless{\section{Exact Natural Gradients and KFAC}\label{sec:nat_kfac}}

In this section, we will show that KFAC is not strongly related to second-order information. We start with a brief review of KFAC, see Appendix~\ref{sec:kfac_derivation} for more details, and then proceed with experiments.

\begin{figure*}[t]
	\vskip -10pt
	\centering
	\includegraphics[width=0.95\textwidth]{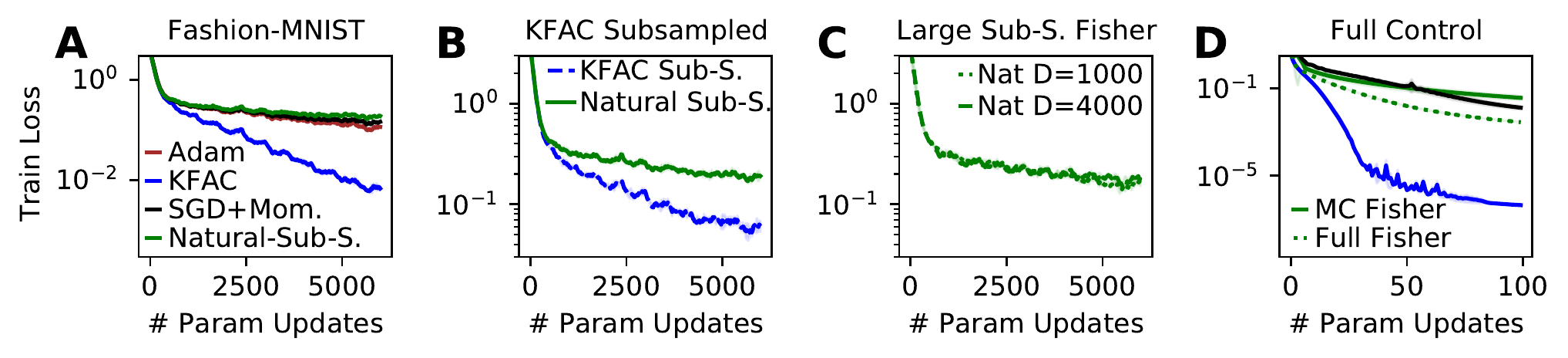}
	\vskip -10pt
	\caption{
		\textbf{Paradoxically, KFAC -- an approximate second-order method -- outperforms exact second-order udpates in standard as well as important control settings.}
		\textbf{(A)}~Comparison between Subsampled Natural Gradients and KFAC. KFAC performs significantly better. Theoretically, its only advantage over the subsampled method is using more data to estimate the curvature. All methods use a batchsize of 100 and are trained for 10 epochs, with hyperparameters tuned individually for each method (here and in all other experiments).
		\textbf{(B)}~Comparison between Subsampled Natural Gradients and Subsampled KFAC. Both algorithms use exactly the same amount of data to estimate the curvature. From a theoretical viewpoint, KFAC should be a strictly worse approximation of second-order updates than the exact subsampled method; nevertheless, it performs significantly better. 
		\textbf{(C)}~Additional control in which the subsampled Fisher is approximated on larger mini-batches.
		\textbf{(D)}~Full control setting, in which the training set is restricted to 1000 images and gradients and curvature are computed on the entire batch (in addition, for a clean comparison KFAC does not use an exponential average to estimate the curvature). The dashed green line corresponds to exact natural gradients without any approximations. 
		Consistent with prior literature, full second-order updates do outperform standard first-order updates (dashed green vs.\ black line). More importantly, and very surprisingly, KFAC significantly outperforms exact second-order updates. This is very strong evidence that KFAC is not closely related to Natural Gradients.\\
		\textbf{(A-D)} We repeat several key experiments with other datasets and architectures and results are consistent with the ones seen here, see main text and appendix. \textbf{(A-D)} Solid lines show mean across three seeds; shaded regions (here and in remaining main paper figures) show mean$\pm$std, but for most experiments are visually hard to distinguish from the mean.
	}
	\vskip -9pt
	\label{fig:kfac_ablation_data}
\end{figure*}

\subsection{Review of KFAC}
KFAC makes two approximations to the Fisher. Firstly, it only considers diagonal blocks of the Fisher, where each block corresponds to one layer of the network. Secondly, each block is approximated as a Kronecker product $(\A\A^T)\otimes (\E_F\E_F^T)$.
This approximation of the Fisher leads to the following update.
\begin{align}\label{eq:kfac_update}
	(\Delta \W)^T &= \left(\A \A^T +\lambda_A \I\right)^{-1}  \left(\A \E^T\right) \left(\E_F \E_F^T +\lambda_E \I\right)^{-1}
\end{align}
where $\lambda_A, \lambda_E$ are damping terms satisfying $\lambda_A \cdot \lambda_E = \lambda$  for a hyperparameter $\lambda$ and ${\frac{\lambda_A}{\lambda_E} = \frac{n\cdot\mathrm{Tr}(\A\A^T)}{m\cdot\mathrm{Tr}(\E_F\E_F^T)}}$.

\textbf{Heurisitc Damping.} We emphasise that the damping performed here is heuristic: Every Kronecker factor is damped individually. This deviates from the theoretically ``correct'' form of damping, which consists of adding a multiple of the identity to the approximate curvature.
To make this concrete, the two strategies use the following damped curvature matrices
\begin{align}
	\text{standard:}\quad& \left(\A\A^T \otimes \E_F\E_F^T\right) + \lambda \I \\
	\text{heuristic:}\quad& \left(\A\A^T +\lambda_A\I\right)\otimes \left(\E_F\E_F^T + \lambda_E \I\right)
\end{align}
Heuristic damping adds undesired cross-terms ${\lambda_E \A\A^T\otimes \I}$ and $\lambda_A \I\otimes\E_F\E_F^T$ to the curvature, and we point out that these cross terms are typically much larger than the desired damping $\lambda \I$.
While the difference in damping may nevertheless seem innocuous,
\citet{martens2015optimizing, ba2016distributed, george2018fast} all explicitly state that heuristic damping performs better than standard damping. From a theoretical perspective, this is a rather mysterious observation.


In practice, the Kronecker factors $\A \A^T$ and $\E_F\E_F^T$ are updated as exponential moving averages, so that they incorporate data from several recent mini-batches. 


\textbf{Subsampled Natural Gradients vs KFAC}: There are two high level differences between KFAC and subsampled natural gradients.
(1) KFAC can use more data to estimate the Fisher, due to its exponential moving averages. (2) For a given mini-batch, natural gradients are exact, while KFAC makes additional approximations.

A priori, it seems that (1) is a disadvantage for subsampled natural gradients, while (2) is an advantage. However, we will see that this is not the case.




\subsection{Experiments}
The first set of experiments is carried out on a fully connected network on Fashion MNIST \cite{xiao2017fashion} and followed by results on a Wide ResNet \cite{he2016deep} on CIFAR10 \cite{krizhevsky09learningmultiple}. 
We run several additional experiments, which are presented fully in the appendix, and will be refered to in the main text. These include repeating the first set of experiments on MNIST; results on CIFAR100; a VGG network \cite{simonyan2014very} trained on SVHN \cite{netzer2011reading} and more traditional autoencoder experiments \cite{hinton2006reducing}.

We emphasise that, while our results are surprising, they are certainly not caused by insufficient hyperparameter tuning or incorrect computations of second-order updates. 
In particular, we perform independent grid searches for each method and ablation and make sure that the grids are sufficiently wide and fine. Details are given in Appendix \ref{sec:exp_details} and \ref{sec:tuning_NG} and we describe part of our software validation in \ref{sec:software_validation}. Code to validate and run the software is provided. 
Moreover, as will be pointed out throughout the text, our results are consistent with many experiments from prior work. 

\begin{figure*}[t]
	\centering
	\includegraphics[width=0.95\textwidth]{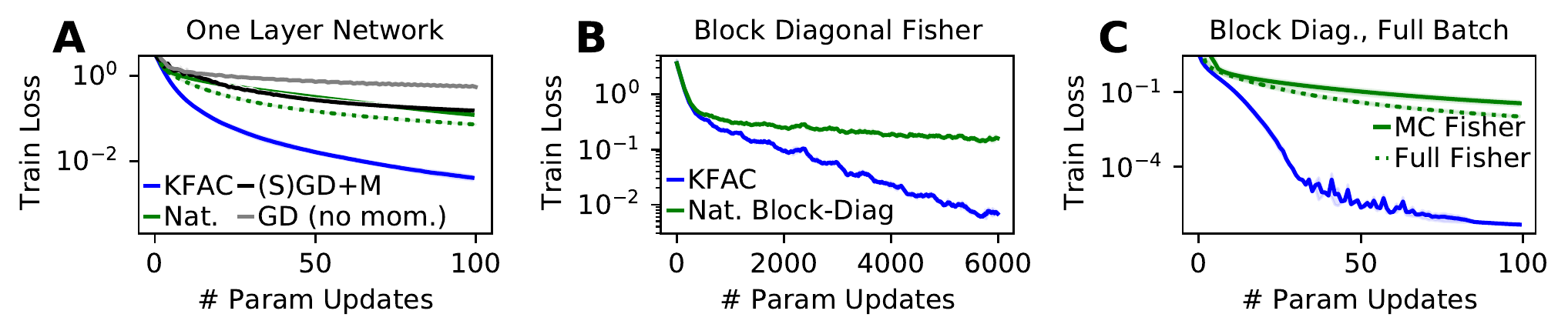}
	\vskip -10pt
	\caption{\textbf{Advantage of KFAC over exact, subsampled Natural Gradients is not due to block-diagonal structure.} 
		\textbf{(A)}~A one layer network (i.e.\ we perform logistic regression) is trained on 1000 images and full batch gradients are used. 
		In particular, KFAC and the subsampled method use the same amount of data to estimate the curvature.		
		In a one layer network the block-diagonal Fisher coincides with the full Fisher, but KFAC still clearly outperforms natural gradients.
		\textbf{(B)}~Comparison between KFAC and layerwise (i.e.\ block-diagonal) subsampled Natural Gradients on full dataset with a three layer network. 	
		\textbf{(C)}~Same as (B), but training set is restricted to a subset of 1000 images and full-batch gradient descent is performed. 
		\textbf{(A-C)}Experiments on Fashion MNIST, results on MNIST are analogous, see appendix.
	}
	\label{fig:kfac_ablation_bd}
	\vskip -3pt
\end{figure*}

To obtain easily interpretable results without unnecessary confounders, we choose a constant step size for all methods, and a constant damping term. 
This matches the setup of prior work \cite{desjardins2015natural, zhang2018three, george2018fast, goldfarb2020practical}. We re-emphasise that these hyperparameters are optimized carefully and indepently for each method and experiment individually.

Following the default choice in the KFAC literature \cite{martens2015optimizing}, we usually use a Monte Carlo estimate of the Fisher, based on sampling one label per input. We will also carry out controls with the Full Fisher.

\textbf{Performance:} We first investigate the performance of KFAC and subsampled natural gradients, see Figure~\ref{fig:kfac_ablation_data}A. Surprisingly, natural gradients significantly underperform KFAC, which reaches an approximately 10-20x lower loss on both Fashion MNIST and MNIST. 
This is a concerning finding, requiring further investigation: After all, the exact natural gradient method should in theory perform at least as good as any approximation of it. Theoretically, the only potential advantage of KFAC over subsampled natural gradients is that it uses more information to estimate the curvature.

\textbf{Controlling for Amount of Data used for the Curvature:} The above directly leads to the hypothesis that KFAC's advantage over subsampled natural gradients is due to using more data for its approximation of the Fisher. To test this hypothesis, we perform three experiments. (1) We explicitly restrict KFAC to use the same amount of data to estimate the curvature as the subsampled method. (2) We allow the subsampled method to use larger mini-batches to estimate the Fisher. (3) We restrict the training set to 1000 (randomly chosen) images and perform full batch gradient descent, again with both KFAC and subsampled natural gradients using the same amount of data to estimate the Fisher. Here, we also include the Full Fisher information as computed on the 1000 training samples, rather than simply sampling one label per datapoint (MC Fisher). In particular, we evaluate exact natural gradients (without any approximations: The gradient is exact, the Fisher is exact and the inversion is exact). 
The results are shown in Figure~\ref{fig:kfac_ablation_data} and all lead to the same conclusion: The fact that KFAC uses more data than subsampled natural gradients does not explain its better performance. In particular, subsampled KFAC outperforms exact natural gradients, also when the latter can be computed without any approximations.

This first finding is very surprising. Nevertheless, we point out that it is consistent with experimental results from prior work as well as commonly held beliefs. Firstly, it is widely believed that subsampling natural gradients leads to poor performance. This belief is partially evidenced by claims from \citet{martens2010deep} and often mentioned in informal discussions and reviews. It matches our findings and in particular Figure~\ref{fig:kfac_ablation_data}D, which shows that benefits of Natural Gradients over SGD only become notable when computing the Fisher fully.\footnote{It also evidences the correctness of our implementation of natural gradients.}
Secondly, we have shown that KFAC performs well even when it is subsampled in the same way as we subsampled natural gradients. While this does seem to contradict the belief that natural gradient methods should not be subsampled, it is confirmed by experiments from \citet{botev2017practical, bernacchia2019exact}: See Figure 2 "per iteration curvature" in \citet{botev2017practical} and note that in \citet{bernacchia2019exact} the curvature is evaluated on individual minibatches.

\textbf{Additional Experiments:} We repeat the key experiments from Figure~\ref{fig:kfac_ablation_data}A,B in several additional settings: On a MLP on MNIST, on a ResNet with and without batch norm on CIFAR10 and for traditional autoencoder experiments. The findings are in line with the ones above, and solidify concerns whether KFAC is related to second-order information.

\begin{figure*}[t]
	\vskip - 6pt
	\centering
	\includegraphics[width=.95\textwidth]{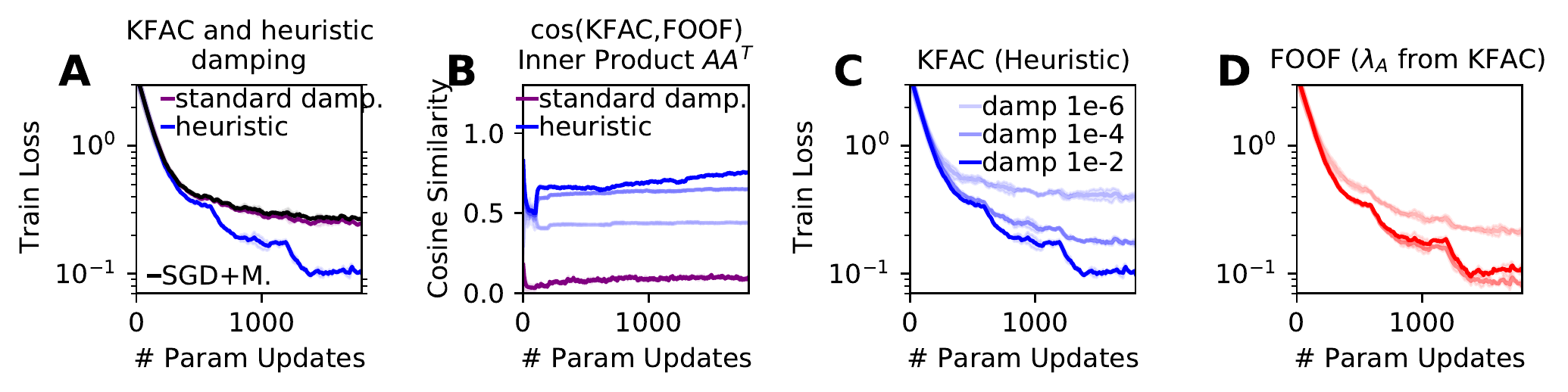}
	\vskip -10pt
	\caption{\textbf{Heuristic Damping increases KFAC's performance as well as its similarity to first-order method FOOF.}
		\textbf{(A)}~Heuristic damping is strictly needed for performance of KFAC; with standard damping, KFAC performs similar to SGD. 
		\textbf{(B)}~Heuristic damping significantly increases similarity of KFAC to FOOF. For the inner product space, we use the ``curvature'' matrix of FOOF.
		\textbf{(C+D)}~Performance of KFAC and FOOF across different damping strengths using heuristic damping for KFAC. For a clean and fair comparison, this version of FOOF uses $\lambda_A$ from KFAC, see Appendix~\ref{sec:foof_lambda_kfac}.
		Notably, FOOF already works well for lower damping terms than KFAC, suggesting that KFAC requires larger damping mainly to guarantee similarity to FOOF and limit the effect of its second Kronecker factor.
		\textbf{(A-D)} and our theoretical analysis suggest that KFAC owes its performance to similarity to the first-order method FOOF.
		Experiments are on Fashion-MNIST, results on MNIST are analogous, see appendix.
		We also re-run experiment (A) in several other settings and confirm that heuristic damping is crucial for performance, see Appendix. This is in line with reports from \cite{martens2015optimizing, ba2016distributed, george2018fast}.
	}
	\label{fig:kfac_vs_first}
	\vskip -9pt
\end{figure*}

\textbf{Controlling for Block-Diagonal Structure:} This begs further investigation into why KFAC outperforms natural gradients. KFAC approximates the Fisher as block-diagonal. 
To test whether this explains KFAC's advantage, we conduct two experiments. First, we train a one layer network on a subset of 1000 images with full-batch gradient descent (i.e. we perform logistic regression).
In this case, the block-diagonal Fisher coincides with the Fisher. So, if the block-diagonal approximation were responsible for KFAC's performance, then for the logistic regression case, natural gradients should perform as well as KFAC or better. However, this is not the case as shown in Figure~\ref{fig:kfac_ablation_bd}A. 
As an additional experiment, we consider a three layer network and approximate the Fisher by its block-diagonal (but without approximating blocks as Kronecker products). The resulting computations and inversions can be carried out efficiently akin to the subsampled natural gradient method.
We run the block-diagonal natural gradient algorithm in two settings: In a minibatch setting, identical to the one shown in Figure~\ref{fig:kfac_ablation_data} and in a full-batch setting, by restricting to a subset of 1000 training images. The results in Figure~\ref{fig:kfac_ablation_bd}B,C confirm our previous findings: (1) KFAC significantly outperforms even exact block-diagonal natural gradients (with full Fisher and full gradients). (2) It is not the block-diagonal structure that explains KFAC's performance.

\textbf{Heuristic Damping.} KFAC also deviates from exact second-order updates through its heuristic damping. To test whether this difference explains KFAC's performance, we implemented a version of KFAC with standard damping.\footnote{This can be done with ideas from \cite{george2018fast}.} 
Figure~\ref{fig:kfac_vs_first}A shows that KFAC owes essentially all of its performance to the damping heuristic. This finding is confirmed by experiments on CIFAR10 with a ResNet and on autoencoder experiments. 
We re-emphasise that KFAC outperforms exact natural gradients, and therefore the damping heuristic cannot be seen as giving a better approximation of second-order updates. Rather, heuristic damping causes performance benefits through some other effect. 

\textbf{Summary.} We have seen that KFAC, despite its motivation as an approximate natural gradient method, behaves very differently from true natural gradients. In particular, and surprisingly, KFAC drastically outperforms natural gradients. Through a set of careful controls, we established that KFAC's advantage relies on a seemingly innocuous damping heuristic, which is unrelated to second-order information. We now turn to why this is the case.



\begin{figure*}[t]
	\vskip -10pt
	\centering
	\includegraphics[width=0.85\textwidth]{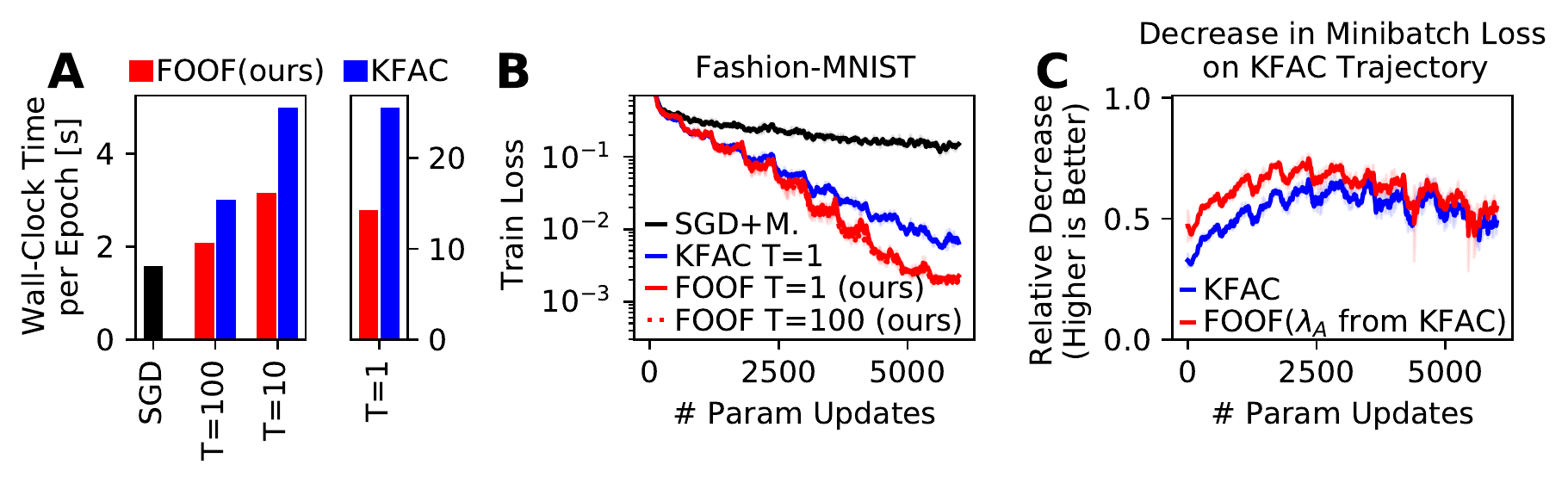}
	\vskip -16pt
	\caption{\textbf{FOOF outperforms KFAC in terms of both per-update progress and computation cost.} 
		\textbf{(A)} Wall-clock time comparison between FOOF, KFAC and SGD. $T$ denotes how frequently matrix inversions (see eq \eqref{eq:foof_update}) are performed. Implemented with PyTorch and run on a GPU. Increasing $T$ above 100 does not notably improve runtime. FOOF is approximately 1.5x faster than KFAC.
		\textbf{(B)}~Training loss on Fashion MNIST. FOOF is more data efficient and stable than KFAC.
		\textbf{(C)}~Comparison of KFAC and a version of KFAC which drops the second Kronecker factor (equivalently, this corresponds to FOOF with damping term $\lambda_A$ from KFAC). We follow the trajectory of KFAC, and at each point we compute the relative improvement of the loss on the current mini-batch that is achieved by the update of KFAC and FOOF, respectively. We use the learning rate and damping that is optimal for KFAC, and scale the FOOF update to have the same norm as the KFAC update at each layer. FOOF performs better, further suggesting that similarity to FOOF is responsible for KFAC's performance. \textbf{(B+C)} See also appendix Fig \ref{fig:faces} for an instance, where FOOF makes more progress per update, but the overall KFAC trajectory performs better. 
	}
	\label{fig:foof}
	\vskip -9pt
\end{figure*}

\tocless{\section{First-order Descent on Neurons}
	\label{sec:foof}}

We will first describe the optimizer "Fast First-Order Optimizer" or "FOOF"\footnote{$F_2O_2$ is a chemical also referred to as ``FOOF''.
} and then explain KFAC's link to it. 

FOOF's update rule is similar to some prior work \cite{desjardins2015natural, frerix2017proximal, amid2021locoprop} and is also related to the idea of optimizing modules of a nested function independently \cite{lecun1988theoretical, carreira2014distributed, taylor2016training, gotmare2018decoupling}
. The view on optimization which underlies FOOF is principled and new, and, among other differences, our insights and experiments linking KFAC to FOOF are new. For a more detailed discussion see Appendix~\ref{sec:related_similar}


Recall our notation for one layer of a neural network from Section~\ref{sec:notation}, namely $\A, \W$ for input activation and weight matrix as well as $\B = \W \A$ and $\E = \frac{\partial L}{\partial B}$. 

Typically, for first-order optimizers, we compute the weights' gradients for each datapoint and average the results.
Changing perspective, we can try to find an update of the weight matrix that explicitly changes the layer's outputs $\B$ into their gradient direction $\E = \frac{\partial L}{\partial \B}$. In other words, we want to find a weight update $\Delta\W$ to the parameters $\W$, so that the layer's output changes in the gradient direction, i.e.\ $(\W + \Delta \W)\A = \B + \eta\frac{\partial L}{\partial \B}$ or equivalently $(\Delta \W)\A = \eta\E$ for a learning rate $\eta$. 
Formally, we optimize
\begin{align}\label{eq:foof_minimise}
	\min_{\Delta \W\in \mathbb{R}^{n\times m}} \Vert (\Delta\W) \A - \eta\E\Vert^2 + \frac{\lambda}{2} \Vert \Delta \W \Vert^2
\end{align}
where the second summand $\frac{\lambda}{2} \Vert\Delta\W\Vert^2$ is a proximity constraint limiting the update size. \eqref{eq:foof_minimise} is a linear regression problem (for each row of $\Delta\W$) solved by
\begin{align}
	\label{eq:foof_update}
	\left(\Delta \W\right)^T = \eta\left(\lambda \I + \A \A^T\right)^{-1}\A \E^T
\end{align}
Pseudocode for the resulting optimizer FOOF is presented in Appendix~\ref{sec:pseudo}. Figures~\ref{fig:foof},\ref{fig:cifar} show the empirical results of FOOF, which outperforms not only SGD and Adam, but often also KFAC. An intuition for why "gradient descent on neurons" performs considerably better than "gradient descent on weights" is that it trades off conflicting gradients from different data points more effectively than the simple averaging scheme of SGD. See Appendix \ref{sec:toy_foof} for an illustratitive toy example for this intuition.

The FOOF udpate can be seen as preconditioning by $\left((\lambda\I + \A\A^T) \otimes \I\right)^{-1}$ and we emphasise that this matrix contains no dependence on the loss, or first/second derivatives of it, so that it cannot be seen as a "second-order" optimizer according to common ML terminology.

\subsection{Stochastic Version of FOOF and Amortisation}
The above formulation is implicitly based on full-batch gradients. 
To apply it in a stochastic setting, we need to take some care to limit the bias of our updates. In particular, for the updates to be completely unbiased one would need to compute $\A \A^T$ for the entire dataset and invert the corresponding matrix at each iteration. This is of course too costly and instead we keep an exponentially moving average of mini-batch estimates of $\A \A^T$, which are computed during the standard forward pass. To amortise the cost of inverting this matrix, we only perform the inversion every $T$ iterations. This leads to slightly stale values of the inverse, but in practice the algorithm is remarkably robust and allows choosing large values of $T$ as also shown in Figure~\ref{fig:foof}.

FOOF can be straightforwardly combined with momentum and (decoupled) weight decay.

\subsection{KFAC as First-Order Descent on Neurons}
Recall that the KFAC update is given by
\begin{align*}
	(\Delta \W)^T &= \left(\A \A^T +\lambda_A \I\right)^{-1}  \left(\A \E^T\right) \left(\E_F \E_F^T +\lambda_E \I\right)^{-1}.
\end{align*}
\begin{figure*}
	\vskip -3pt
	\centering
	\includegraphics[width=0.95\textwidth]{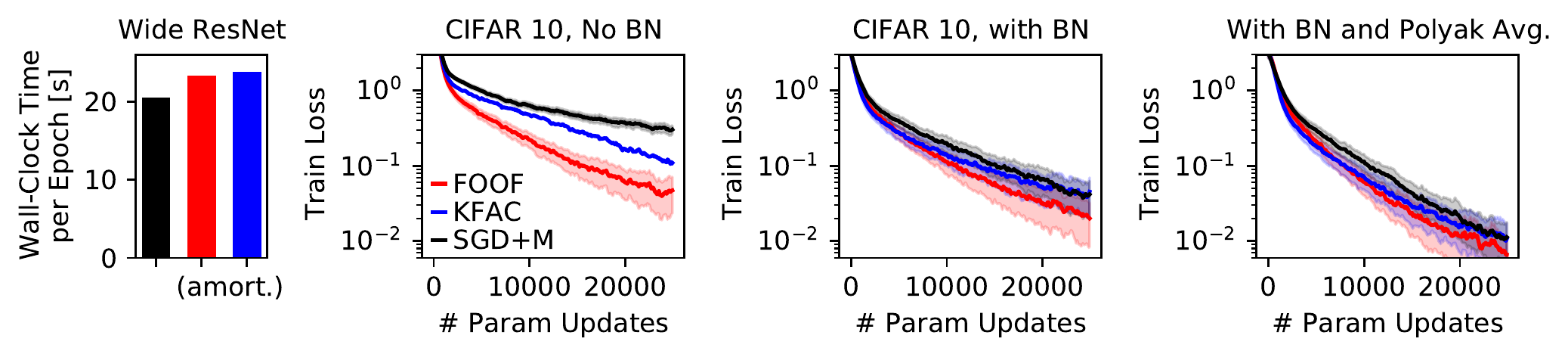}
	\vskip -10pt
	\caption{\textbf{FOOF outperforms KFAC in a Wide ResNet18 on CIFAR 10.} 
		\textbf{(A)} Wall-clock time comparison between SGD and amortised versions of FOOF, KFAC. In convolutional architectures, FOOF and KFAC can be effectively amortised wihtout sacrificing performance. 
		\textbf{(B, C, D)} Training loss in different settings. \textbf{(A-D)} Results on CIFAR100 and SVHN are analogous.
	}
	\label{fig:cifar}
	\vskip -9pt
\end{figure*}
\textbf{Similarity of KFAC to FOOF and Damping:} The update of KFAC differs from the FOOF update (eq~\eqref{eq:foof_update}) only through the second factor 
$\left(\E_F \E_F^T +\lambda_E \I\right)^{-1}$. We emphasise that this similarity is induced mainly through the heuristic damping strategy. In particular, with standard damping, or without damping,
the second Kronecker factor of KFAC could lead to updates that are essentially uncorrelated with FOOF. However, as we use heuristic damping and increase the damping strength $\lambda_E$, the second factor will be closer to (a multiple of) the identity and KFAC's update will become more and more aligned with FOOF. \\
Based on this derivation, we now test empirically whether heuristic damping indeed makes KFAC similar to FOOF and how it affects performance.
\\
Figure~\ref{fig:kfac_vs_first}B confirms our theoretical argument that heuristic damping drastically increases similarity of KFAC to FOOF and stronger heuristic damping leads to even stronger similarity. This similarity is directly linked to performance of KFAC as shown in Figure~\ref{fig:kfac_vs_first}C.
These findings, in particular the necessity to use heuristic damping, already strongly suggest that similarity to FOOF is required for KFAC to perform well.
Moreover, as shown in Figure~\ref{fig:kfac_vs_first}D, FOOF requires lower damping than KFAC to perform well. This further suggests that damping in KFAC is strictly required to limit the effect of $\E_F \E_F^T$ on the update, thus increasing similarity to FOOF. 
All in all, these results directly support the claim that KFAC, rather than being a natural gradient method, owes its performance to approximating FOOF.

\textbf{Performance:} If the above view of KFAC is correct, and it owes its performance to similarity to FOOF, 
then one would expect FOOF to perform better than or similarly to KFAC. 
We carry out two different tests of this hypothesis. First, we train a network using KFAC and at each iteration, we record the progress KFAC makes on the given mini-batch, measured as the relative decrease in loss. We compare this to the progress that KFAC would have made without its second Kronecker factor. We use learning rate and damping that are optimal for KFAC and, when dropping the second factor, we rescale the update to have the same norm as the original KFAC update for each layer. The results are shown in Figure~\ref{fig:foof}C and show that without the second factor, KFAC makes equal or more progress, which supports our hypothesis. 
This observation is consistent across all different experimental setups we investigated, see appendix. 
\\
As a second test, we check whether FOOF outperforms KFAC, when both algorithms follow their own trajectory. This is indeed the case as shown in Figure~\ref{fig:foof}B. The only case where the advantage described in Figure~\ref{fig:foof}C does not translate to an overall better performance is the autoencoder setting, as analysed in the appendix, e.g.\ Figure \ref{fig:faces}, Sec \ref{sec:additional_mnist}.
Results on a Wide ResNet18 demonstrate that our findings carry over to more complex settings and that FOOF often outperforms KFAC, see Figure~\ref{fig:cifar}.

\textbf{Computational Cost:}
We also note that, on top of making more progress per parameter update, FOOF requires strictly less computation than KFAC: It does not require an additional backward pass to estimate the Fisher; it only requires keeping track of, inverting as well as multiplying the gradients by one matrix rather than two (only $\A\A^T$ and not $\E_F\E_F^T$). These savings lead to a 1.5x speed-up in wall-clock time per-update for the amortised versions of KFAC and FOOF as shown in Figure~\ref{fig:foof}A.

\textbf{Cost in Convolutional Architectures:} The only overhead of KFAC and FOOF which cannot be amortised is performing the matrix multiplications in eqs~\eqref{eq:foof_update},\eqref{eq:kfac_update}. These are standard matrix-matrix multiplications and are considerably cheaper than convolutions, so that we found that KFAC and FOOF can be amortised to have almost the same wall-clock time per update as SGD for this experiment ($\sim$10\% increase for FOOF, $\sim$15\% increase for KFAC) without sacrificing performance, see Appendix~\ref{sec:exp_details}. We note that these results are significantly better than wall-clock times from \citet{ba2016distributed, desjardins2015natural}, which require approximately twice as much time per update as SGD.\footnote{Information reconstructed from Figure 3 in \citet{ba2016distributed} and Figure 4 in \citet{desjardins2015natural}.}

\textbf{Summary:} 
We had already seen that KFAC does not rely on second-order information. In addition, these results suggest that KFAC owes its strong performance to its similarity to FOOF, a principled, well-performing first-order optimizer.

\tocless{\section{Limitations}}
In our experiments, we report training losses and tune hyperparameters with respect to them. While this is the correct way to test our hypotheses and common for developing and testing optimizers \cite{sutskever2013importance}, it will be important to test how well the optimizers investigated here generalise. A meaningful investigation of generalisation requires a different experimental setup (as demonstrated in \citet{zhang2018three}) and is left to future work.  With this in mind, we note that in our setting the advantage of FOOF and KFAC in training loss typically translates to an advantage in validation accuracy (and that FOOF and KFAC behave similarly). 

We have restricted our investigation to the context of optimization and more specifically KFAC. While we strongly believe that our findings carry over to other Kronecker-factored optimizers \cite{desjardins2015natural, goldfarb2020practical, botev2017practical, george2018fast, martens2018kronecker, osawa2019large}, we have not explicitly tested this. 

While, in all our experiments, the newly proposed view that KFAC is closely related to FOOF captures considerably more characteristics of KFAC than the standard view of KFAC as a natural gradient method, we highlight again that there are some limitations to this explanation and, in particular, one setting where KFAC performs slightly better than FOOF, see Section \ref{sec:additional_mnist}.

\tocless{\section{Discussion}
	\label{sec:discussion}}
The purpose of this discussion is twofold. On the one hand, we will show that, while being surprising and contradicting common, strongly held beliefs, much of our results are consistent with data from prior work. 
On the other hand, we will summarise in how far our fundamentally new explanation for KFAC's effectiveness improves upon prior knowledge and resolves several puzzling observations.

\textbf{Natural Gradients vs KFAC:} Our first key result is that KFAC outperforms exact natural gradients, despite being motivated as an approximate natural gradient method. 
We perform several controls and a particularly important set of experiments is comparing exact, subsampled natural gradients to subsampled KFAC in a range of settings. In these experiments, we find: (1) Subsampled natural gradients often do not perform much better than SGD with momentum. (2) Subsampled KFAC works very well.
Finding (1) is consistent with rather common beliefs that subsampling the curvature is harmful. These beliefs are often uttered in informal discussions and are partially evidenced by claims from \citet{martens2010deep}. Moreover, in some controls (e.g.\ Fig~\ref{fig:kfac_ablation_data}D), we show that full (non-subsampled) natural gradients do outperform SGD with momentum, consistent for example with \citet{martens2010deep}.
Thus, finding (1) is in line with prior knowledge and results.
Moreoever, finding (2) matches experiments from \citet{botev2017practical, bernacchia2019exact} as described in Section~\ref{sec:nat_kfac} and thus also is in line with prior work.

Also independently of our results, it is worth noting that the performance of subsampled KFAC reported in \citet{botev2017practical, bernacchia2019exact} is hard to reconcile with the simultaneous convictions that (1) KFAC is a natural gradient method and (2) subsampling the Fisher has detrimental effects.

Our newly suggested explanation of KFACs performance resolves this contradiction.
Even if one were to disagree with our explanation for KFAC's effectiveness, the above is an important insight, strengthened by our careful control experiments, and deserves further attention.

It is also worth noting that another natural way to check if our finding that KFAC outperforms Natural Gradients agrees with prior work would be to look for a direct comparison of KFAC with Hessian-Free optimization (HF). 
Perhaps surprisingly, to the best of our knowledge, there is no meaningful comparison between these two algorithms in the literature.\footnote{We are aware of only one comparison, which unfortunately has serious limitations, see Appendix~\ref{sec:hf_vs_kfac}.}
It will be interesting to see a thoroughly controlled, well tuned comparison between HF and KFAC.

\textbf{Damping:} A second cornerstone of our study is the effect of damping on KFAC.
We found that employing a heuristic, rather than standard damping strategy is essential for performance. The result that heuristic damping improves KFAC's performance has been noted several times (qualitatively, rather than quantitatively), see the original KFAC paper \cite{martens2015optimizing}, its large-scale follow up \cite{ba2016distributed}, and even E-KFAC \cite{george2018fast}.

While choice of damping strategy may seem like a negligible detail at first, it is important to bear in mind that without heuristic damping KFAC performs like standard first-order optimizers like SGD or Adam. Thus, if we want to understand KFAC's effectiveness, we have to account for its damping strategy. This is achieved by our new explanation and even if one were to disagree with it, this finding deserves further attention and an explanation. 

\textbf{Ignoring the second Kronecker factor $\E_F \E_F^T$ of the approximate Fisher:} A third important finding is that KFAC performs well, and often better, without its second Kronecker factor. The fact that algorithms that are similar to KFAC without the second factor perform exceptionally well is consistent with prior work \cite{desjardins2015natural, frerix2017proximal, amid2021locoprop}. 
Moreover, \citet{desjardins2015natural} explicitly state that their algorithm performs more stably without the second Kronecker factor, which further confirms our findings. 
We emphasise that, without the second Kronecker factor, the preconditioning matrix of KFAC is independent of the loss (or derivatives of it), and thus cannot be seen as a classical second-order method. 
From a second-order viewpoint, dropping dependence on the loss should have detrimental effects, inconsistent with results from the studies above as well as ours. 

\textbf{Architectures with Parameter Sharing:} Finally, we point to another intriguing finding from \citet{grosse2016kronecker}. For architectures with parameter sharing, like CNNs or RNNs, approximating the Fisher by a Kronecker product requires additional, sometimes complex assumptions, which are not always satisfied. \citet{grosse2016kronecker} explicitly investigate one such assumption for CNNs, pointing out that it is violated in architectures with average- rather than max-pooling. Nevertheless, KFAC performs very well in such architectures \cite{ba2016distributed, george2018fast, zhang2018three, osawa2019large}. 
This suggests that KFAC works well independently of how closely it is related to the Fisher, which is a puzzling observation when viewing KFAC as a natural gradient method. Again, our new explanation resolves this issue, since KFAC still performs gradient descent on neurons.\footnote{In eq \eqref{eq:foof_minimise}, we now impose one constraint per datapoint and per ``location'' at which the parameters are applied.}

In summary, we have shown that viewing KFAC as a second-order, natural gradient method is irreconcilable with a host of experimental results, from our as well as other studies. We then proposed a new, considerably improved explanation for KFAC's effectiveness. We also showed that the algorithm FOOF, which results from our explanation, can give further performance improvements compared to the state-of-the art optimizer KFAC.

\section*{Acknowledgments}
I thank James Martens, Laurence Aitchison and Yann Ollivier for very helpful feedback on earlier versions of this manuscript. I am also thankful for many helpful discussions with Laurence Aitchison, Robert Meier, Asier Mujika and Kalina Petrova leading to the conception and development of this project.

\bibliographystyle{apalike}
\bibliography{sample}

\clearpage

\onecolumn
\appendix
\setcounter{tocdepth}{4}
\tableofcontents
\begin{enumerate}
	\item[\ref{sec:tuning_NG}] \textbf{Tuning Natural Gradients}
	\item[\ref{sec:software_validation}] \textbf{Software Validation}
	\item[\ref{sec:applications}] \textbf{Further Applications of Implicit, Fast Fisher Inversion}
	\item[\ref{sec:exp_details}] \textbf{Experimental Details (incl.\ HP tuning and values)}
	\item[\ref{sec:kfac_derivation}] \textbf{Derivation of KFAC}
	\item[\ref{sec:toy_foof}] \textbf{Toy Example Illustrating the Difference between SGD and FOOF}
	\item[\ref{sec:kfac_laplace}] \textbf{Kronecker-Factored Curvature Approximations for Laplace Posteriors}
	\item[\ref{sec:related}] \textbf{Related Work}
	\item[\ref{sec:nat_details}] \textbf{Details for Efficiently Computing $\F^{-1}$-vector products for a Subsampled Fisher}
	\item[\ref{sec:additional_experiments}] \textbf{Additional Experiments}
\end{enumerate}
\renewcommand\thefigure{\thesection.\arabic{figure}}

\section{Tuning Natural Gradients}
\label{sec:tuning_NG}
While, for the sake of fairness, all results in the paper are based on the hyperparameter optimization scheme described further below, we emphasise that we also invested effort into hand-tuning subsampled natural gradient methods, but that this did not give notably better results than the ones reported in the paper. For our Fashion MNIST and MNIST experiments, we also implemented and tested an adaptive damping scheme as described in \cite{martens2015optimizing}, tried combining it with automatic step size selection and also a form of momentum, all as described in \cite{martens2015optimizing}. None of these techniques gave large improvements. 

\section{Software Validation}
\label{sec:software_validation}
To validate that our algorithm of computing products between the damped, inverse Fisher and vectors (as described in Appendix~\ref{sec:nat_details}) is correct, we considered small networks in which we could explicitly compute and invert the Fisher Information and confirmed that our implicit calculations agree with the explicit calculations for both fully connected as well as convolutional architectures.

Moreover, the fact that natural gradients outperform SGD in Figure~\ref{fig:kfac_ablation_data}D (and some other settings) is very strong evidence that our implementation of natural gradients is correct.

\section{Further Applications of Implicit, Fast Fisher Inversion}\label{sec:applications}
\subsection{Bayesian Laplace Posterior Approximation}
Laplace approximations are a common approximation to posterior weight distributions and various techniques have been proposed to approximate them, for a recent overview and evaluation we refer to \citet{daxberger2021laplace}. In this context, the Hessian is the posterior precision matrix and it is often approximated by the Fisher or empirical Fisher, since these are positive semi-definite by construction. 

It is often stated that sampling from a full covariance, Laplace posterior is computationally intractable. However, combining our insights with an additional trick allows us to sample from an exact posterior, given a subsampled Fisher, as shown below. We adapted the trick from \citet{thanks_arnaud}, who credits \citet{hoffman1991constrained}.
We also remark that \citet{immer2021scalable} sample from the predictive distribution of a linearised model, arguing theoretically and empirically that the linearised model is the more principled choice when approximating the Hessian by the Fisher.

\subsubsection{Sampling from the full covariance Laplace posterior}
We write $\priorprec$ for the prior precision and $\La = \priorprec + D\F$ for the posterior precision, where $\F$ is the Fisher. 
All expressions below can be evaluated efficiently for example if the prior precision is diagonal and constant across layers, as is usually the case. 
As for the natural gradients, we factorise $\F = \G\G^T$, where $\G$ is a $N\times D$ matrix ($N$ is the number of parameters, $D$ the number of datapoints).
Using Woodbury's identity, we can write the posterior variance as

\begin{align}
	\La^{-1} = \left(\priorprec + D\G \G^T\right)^{-1} = \priorprec^{-1} - D\priorprec^{-1}\G\left(\I + D\G^T \priorprec^{-1}\G\right)^{-1}\G^T\priorprec^{-1}
\end{align}
We now show how to obtain a sample from this posterior. To this end, define matrices $\V, \U$ as follows:
\begin{align}
	\V &= \left(D^{1/2}\priorprec^{-1/2}\right)\G \\
	\U &= \I + \V^T\V = \I + D\G^T\priorprec^{-1}\G\\
\end{align}
Let $\y\sim\N \left(\0,\I_N\right)$ and $\z\sim\N\left(\0,\I_D\right)$, and define $\x$
\begin{align}
	\x = \y - \V \U^{-1} \left(\V^T \y + \z\right)
\end{align}
We will confirm by calculation that $\priorprec^{-1/2}\x$ is a sample from the full covariance posterior.

$\x$ clearly has zero mean. 
The covariance $\EE\left[\x\x^T\right]$ can be computed as
\begin{align}
	\EE\left[\x\x^T\right] &= \I + \V \U^{-1}\left(\V^T \V + \I\right)\U^{-T}\V^T 
	- 2\V \V^T
\end{align}
Since $\U$ is symmetric and since we chose $\V^T\V + \I = \U$ , the above simplifies to 
\begin{align}
	\EE\left[\x\x^T\right] &= \I - \V\U^{-1}\V^T
\end{align}
By our choice of $\U, \V$, this expression equals
\begin{align}
	\EE\left[\x\x^T\right] &= \I - D\priorprec^{-1/2}\G\left(\I + D\G^T\priorprec^{-1}\G\right)^{-1}\G^T\priorprec^{-1/2}
\end{align}
In other words, $\priorprec^{-1/2}\x$ is a sample form the full covariance posterior.

\subsubsection{Efficient Evaluation of the above procedure}
The computational bottlenecks are computing $\G^T \priorprec^{-1} \G^T$, calculating vector products with $\G$ and $\G^T$. 

Note that for a subsampled Fisher with moderate $D$, we can invert $\U$ explicitly.

We have already encountered all these bottlenecks in the context of natural gradients and they can be solved efficiently in the same way, see Section \ref{sec:nat_details}. The only modification is multiplication by $\priorprec^{-1}$ and for any diagonal prior, this can be solved easily.

\subsection{Continual Learning}
Closely related to Bayesian posteriors is a number of continual learning algorithms \cite{kirkpatrick2017overcoming, nguyen2017variational, benzing2020unifying}. For example, EWC \cite{kirkpatrick2017overcoming} relies on the Fisher to approximate posteriors. Formally, it only requires evaluating products of the form $\v^T \F \v$. Since the Fisher is large, EWC uses a diagonal approximation. 
From the exposition below, it is not to difficult to see that $\v^T \F \v$ is easy to evaluate for a subsampled Fisher and the memory cost is roughly equal to that used for a standard for- and backward pass through the model: It scales as the product of the number of datapoints and the number of neurons (rather than weights).

\subsection{Meta Learning / Bilevel optimization}
Some bilevel optimization algorithms rely on the Implicit Function Theorem to estimate outer-loop gradients after the inner loop has converged. They require evaluating a product of the form $(\lambda\I + \H)^{-1} \v$, where $\H$ is the  Hessian and $\v$ a vector, see e.g.\ \cite{bengio2000gradient}. If we approximate the Hessian by a subsampled Fisher, the method used here is directly applicable to compute this product. 

\section{Experimental Details}
\label{sec:exp_details}
All experiments were implemented in PyTorch \cite{paszke2019pytorch}.\footnote{So long and thanks for all the hooks.}
All models use Kaiming-He initialisation \cite{he2015delving, glorot2010understanding}.
Moreover, we average all results across three different random seeds.

\subsection{Exponentially Moving Averages and Subsampled KFAC}
As mentioned in the main text, we use exponentially moving averages to estimate the matrices $\A\A^T$ and $\E_F \E_F^T$. For a quantity $x_t$ the exponentially moving average $\hat{x}_t$ is defined as:
\[
\hat{x}_{t+1} = m \cdot \hat{x}_{t} + (1-m) x_{t+1}
\]
We also normalise our exponentially moving averages (or equivalently initialise $\hat{x}_1 = x_1$). Following \cite{martens2015optimizing}, we set $m=0.95$. Preliminary experiments with $m=0.999$ showed very similar performance.

For subsampled KFAC, we simply set $m=0$, implying that KFAC, like Subsampled Natural Gradients, only uses one mini-batch to estimate the curvature.

\subsection{Hyperparameter Tuning}
Learning rates for all methods were tuned by a grid search, considering values of the form ${1\cdot 10^{i}, 3\cdot 10^{i}}$ for suitable (usually negative) integers $i$. 

The damping terms for Natural Gradients, KFAC, FOOF were determined by a grid searcher over $10^{-6}, 10^{-4}, 10^{-2}, 10^{0}, 10^{2}, 10^{4}, 10^{6}$ on Fashion MNIST and MNIST. We also confirmed that refining the grid to values of the form $10^{i}$ does not meaningfully change results. For all other experiments we used a grid with values of the form $10^{i}$, but also there a coarser grid with values $10^{2i}$ would have given very similar results. 

We extended grids if the performance at a boundary of the grid was optimal (or near optimal).

As already described in Appendix~\ref{sec:tuning_NG}, we did spend additional efforts to tune the natural gradient method, including hand-tuning, as well as using additional ingredients like a fancy, second-order form of momentum, automatic learning rates, adaptive damping, see Appendix~\ref{sec:tuning_NG}.

For SGD, momentum was grid-searched from $0.0, 0.9$. For Adam, we kept most hyperparameters fixed and tuned only learning rate.

Each ablation / experiment got its own hyperparameter search. The only exception is {FOOF~($T=100$)}, which uses exactly the same hyperparameters as FOOF~($T=1$). In particular, for plots like Figure~\ref{fig:kfac_vs_first}C, the learning rate for each damping term was tuned individually. 

We always chose the hyperparameters which gives best training loss at the end of training. Usually, these hyperparameters also outperform others in the early training stage. We also note that several hyperparametrisations of KFAC became unstable towards the end/middle of training, while this did not occur for FOOF.

For the experiments in the main paper, HP values found by this procedure are given in Section \ref{sec:HP_values}.

\subsection{Hyperparameter Robustness}
On top of the learning rate, FOOF requires additional hyperparameters, most of which seem very robust as described below. Additional hyperparameters always include damping and further may include the momentum for exponentially moving averages, and also may include amortisation hyperparameters S,T described in Algorithm~\ref{alg:foof}. 
\begin{itemize}
	\item Step size: Tuning the stepsize of FOOF was as easy/difficult as tuning the step size of SGD in our experience. In particular, BatchNorm allows using a wider range of learning rates without large changes in performance. We recommend to parametrise the stepsize as $\alpha/\lambda$ (where $\lambda$ is the damping term) and (grid-)search over $\alpha$. 
	\item Damping $\lambda$: We searched from a grid of the form $10^{2i}$ for integers $i$ and this was sufficient for our experimental setup. Refining this grid only gave very small improvements (at least for the experimental setup in our paper). Using values that differed by a factor of 100 from the optimal value, usually still gave good results and clearly outperformed SGD.  
	\item Exponential Moving Average $m$ for $\A\A^T$: Following \cite{martens2015optimizing}, we set this value to 0.95. Preliminary experiments with 0.999 gave similar performance. Results seem very robust with respect to this choice. It is conceivable that very small batch sizes require slightly larger values of $m$ to estimate $\A\A^T$ 
	\item Amortisation parameters $S,T$: In our experiments it was sufficient to perform inversions once per epoch. Setting $S$ to $50$ is a mathematically save choice (given $m=0.95$), and setting $S=10$ empirically did not decrease performance either.
\end{itemize}

\subsection{Hyperparameter values}\label{sec:HP_values}
Note that the damping term required for KFAC/FOOF/Natural Gradients depends on implementation details, in particular it depends on the scaling of the curvature matrix (e.g.\ one may or may not normalise by the batch dimension -- there are reasonable arguments for both choices; and in addition, the scaling of the Fisher often varies depending on definition/context/implementation).  
\\
Thus independent implementations may well require different HP values. 

Note that the implementation we used to generate the data for the paper (but not the public code base), scales Fisher differently for MC- and Full Fisher computations, so that damping terms there are not directly comparable.

We also note that for the above reason (and the precise damping strategy employed by KFAC), the damping terms of KFAC and FOOF are not directly comparable. We found that the damping that is applied to the matrix $\A \A^T$ for good HP values of KFAC and FOOF is usually very similar, in line with our overall findings. 


HP values for experiments from main paper in Table \ref{tab:HP_values}.

\begin{center}
	\begin{table}
		\begin{tabular}{l |c|c|c| c|c|c} 
			& SGD+M (lr) & Adam (lr) & Natural MC (lr; damp) & Natural Full & KFAC  & FOOF \\ 
			\hline
			Fig \ref{fig:kfac_ablation_data}A & 0.01 & 1e-4&  1e-5,1e-4& - & 0.3, 1.0& - \\
			Fig \ref{fig:kfac_ablation_data}B & - & -&  same as \ref{fig:kfac_ablation_data}A& - & 0.3, 1.0& - \\
			Fig \ref{fig:kfac_ablation_data}C & - & -&  \makecell{D=1000: 3e-4, 0.01; \\ D=4000: {1e-3, 0.01}}  & - & -& - \\
			Fig \ref{fig:kfac_ablation_data}D &0.03  & &0.1, 1.0 & 100, 100 & 0.1, 1e-4 & - \\
			&&&&&&
			\\
			Fig \ref{fig:kfac_ablation_bd}A & \makecell{0.03 \\ 
				No Mom.: 0.03} & - & 0.01, 0.01 & 100, 100 & 100, 1 & -
			\\
			Fig \ref{fig:kfac_ablation_bd}B& - & - &  Block-Diag: 10, 100&-& same as \ref{fig:kfac_ablation_data}A & -
			\\
			Fig \ref{fig:kfac_ablation_bd}C& - & - &  Block-Diag: 0.1, 1& Block-Diag: 30, 100& 0.1, 1e-4 & -
			\\
			&&&&&&
			\\
			Fig \ref{fig:kfac_vs_first} A & - & - &  -& -& \makecell{std.\ damp: \\ 0.1, 1.0} & -
			\\
			Fig \ref{fig:kfac_vs_first}C+D  & - & - &  -& - &\makecell{1e-6, 1e-6 \\1e-4, 1e-4 \\1e-2, 1e-2 } & \makecell{1e-6, 1e-6 \\1e-4, 1e-4 \\1e-2, 1e-2 \\ see Sec \ref{sec:foof_lambda_kfac}}
			\\
			&&&&&&
			\\
			\makecell{Fig \ref{fig:foof} B \\ (see Fig \ref{fig:kfac_ablation_data}A)}  & - & - &  -& -& - & 30, 100
			
			\\
			&&&&&&
			\\
			Fig \ref{fig:cifar}B    0.003 & - & - &  -& -& 0.03, 1.0 & 10, 100\\
			Fig \ref{fig:cifar}C & 0.03 & - &  -& -& 1.0, 1.0 & 30, 100\\
			Fig \ref{fig:cifar}D & 0.03 & - &  -& -& 1.0, 1.0 & 30, 100
			
		\end{tabular}
		\label{tab:HP_values}
		\caption{HP values for experiments in main paper. Please refer to full text \ref{sec:HP_values} for some notes of caution for interpreting these values. When an entry says "same as ..." this means that the HP was not re-tuned for this figure (we only did this in cases where it makes sense). If two entries are the same, but there is no "same as ...", this means the HP values were tuned independently, and happened to be the same.}
	\end{table}
\end{center}

\subsection{Remaining details}
Experiments were carried out on fully connected networks, with 3 hidden layers of size 1000. The only exception is the network in Figure~\ref{fig:kfac_ablation_bd}A, which has no hidden layer. For simplicity of implementation, we omitted biases. Unless noted otherwise, we trained networks for 10 epochs which batch size 100 on MNIST or Fashion MNIST.
MNIST was preprocessed to have zero mean and unit variance, and -- due to an oversight -- Fashion MNIST was preprocessed in the same way, using the mean+std-dev of MNIST. Our comparisons remain meaningful, as this affects all methods equally and perhaps it even makes our results more comparable to prior work.
Several experiments were carried out on subsets of the training set consisting of 1000 images, in order to make full batch gradient evaluation cheaper and were trained for 100 epochs.

For the wall-clock time experiments, we used PyTorch DataLoaders and optimized the ``pin\_memory'' and ``num\_workers'' arguments for each method/setting.

\subsection{CIFAR 10}
For CIFAR10 we use a Wide ResNet18 and standard data-augmentation consisting of random horizontal flips as well as padding with 4 pixels followed by random cropping. The RGB channels are normalised to have zero mean and unit variance. 

When experimenting with the ResNet with batchnorm, we found that KFAC was unstable with standard momentum. To fix this, we switched on the momentum term only after the first epoch was completed. This seemed helped all methods a bit, so we used it for all of them. We used the same setting for the ResNet without batchnorm. 

When applicable, the SGD baseline uses standard batch norm \cite{ioffe2015batch}. For FOOF and KFAC we include batch norm parameters, but do not train them.

Networks are trained for 50 epochs with batchsize 100. All methods use momentum of 0.9 unless noted otherwise.

For Polyak Averaging, we follow \cite{grosse2016kronecker} and also use the decay value recommended there without further tuning.

\textbf{Amortisation:} We found that KFAC and FOOF can be amortised fairly strongly without sacrificing performance. We invert the Kronecker factors every $T = 500$ timesteps (i.e.\ once per epoch) and only update the exponentially moving averages for the Kronecker factors for the $S=10$ steps immediately before inversion. We also performed thorough hyperparameter searches with $T=250, S=50$ (larger values of $S$ make little sense, due to their limited influence on the exponentially moving averages) which gave essentially identical results. Preliminary experiments with $T=100$ also did not perform better than the schedule described above and used for our experiments.

\subsection{Autoencoder Experiments}
We tried to replicate the setup of \cite{martens2015optimizing} implemented in the tensorflow kfac repository\footnote{\url{https://github.com/tensorflow/kfac/blob/master/kfac/examples/autoencoder\_mnist.py}}. 
Concretly, we use a fully connected network with layers of sizes (d, 1000, 500, 250, 30, 250, 500, 1000, d), where $d$ is the number of input pixels of a single image (which depends on the dataset).
All \textit{hidden} layers, expect for the middle one (of size 30) are followed by a tanh-nonlinearity. 
We preprocessed each dataset to have zero mean and unit variance. We used a batch size of 1000, since \cite{martens2015optimizing} state that small batch sizes lead to too much noise. 
For MNIST and CURVES we trained for 200 epochs, for FACES for 100 epochs. To amortise the runtime, we invert matrices every 10 steps.

\subsection{SVHN}
We used the same preprocessing for SVHN as for CIFAR10. To amortise runtime we used T=100, S=50 (see pseudocode \ref{alg:foof}). We confirmed that with T=S=10 results are essentially the same, suggesting that the amortisation did not harm performance. We used a VGG11 network (without batch norm).

\subsection{Version of FOOF with $\lambda_A$ from KFAC}
\label{sec:foof_lambda_kfac}
In Figures~\ref{fig:kfac_vs_first},\ref{fig:foof}, we use a version of FOOF which always uses exactly the same damping term $\lambda_A$ for each layer as KFAC to obtain a comparison that's as clean as possible.
Concretely, this means carrying out most computations as in KFAC and computing $\E_F\E_F^T$ and $\lambda_A, \lambda_E$ as in KFAC and then computing the parameter update only using the first kronecker factor $\A\A^T$, while omitting the second factor.
Note that $\lambda_A$ varies during training, and typically increases so that this version of FOOF is slightly different from standard FOOF.

Moreover, omitting the second factor notably changes the update size. To correct for this effect, we made the following modifiaction.

In Figure~\ref{fig:foof}C, we rescaled the resulting update (as also described in the figure caption) so that at each layer, the norm of the update had the same norm as the KFAC update.

In the experiment of Figure~\ref{fig:kfac_vs_first}, which we did first, we used a slightly different strategy (but we do not think that this makes a difference).
An increase in $\lambda_A$ leads to a decrease in effective step-size of the above version of FOOF. In KFAC this is compensated by a decrease in damping of the second factor so that the effective stepsize is roughly constant. To compensate analogously in FOOF, we simply multiply the update by the scalar $\lambda_E^{-1}=\lambda_A/\lambda$ to maintain a constant stepsize. For standard FOOF (in all other figures) we use constant $\lambda_A$ and make no such modifications.

\section{Derivation of KFAC}\label{sec:kfac_derivation}
Here, we re-derive KFAC \cite{martens2015optimizing}.

We focus on one layer and use previous notation. In particular, we consider one datapoint $(X_i,y_i)$ (remember that $y_i$ is a sample from the model distribution) and denote the input of the layer as 
$\a_i$, the output as $\b_i=\W\a_i$ and $\e_i = \frac{\partial L}{\partial \b_i}$. For this single datapoint, the block-diagonal part of the Fisher given by this datapoint is exactly equal to 
\[
\F_i = (\a_i\a_i^T) \otimes (\e_i\e_i^T)
\] 
Recall that the Fisher is defined as an expectation over datapoints $\F = \EE[\F_i]$.
If we use a Monte-Carlo approximation of this expectation, the Fisher is approximated as
\[
\F = \sum_{i\in I}\left(\a_i\a_i^T \otimes \e_i\e_i^T\right)
\] 
Now, KFAC makes the following further approximation:
\[
\F = \sum_{i\in I}\left(\a_i\a_i^T \otimes \e_i\e_i^T\right)  \approx
\left(\sum_{i\in I}\a_i\a_i^T\right) \otimes \left(\sum_{i\in I}\e_i\e_i^T\right)
\] 

In general, this approximation is imprecise: It is equivalent to approximating a rank $|I|$ matrix by a rank 1 matrix\footnote{To see this, simply
	reshape the matrix $F$ by first flattening $\a_i \a_i^T, \e_i \e_i^T$ and then replacing the Kronecker product by the standard outer product of vectors. The resulting matrix has the same entries as $F$ and each $F_i$ has rank 1. But in general, the sum over $F_i$ has rank $|I|$.}.
If we take for example the MNIST dataset, which has 60,000 inputs with 10 labels each, the full diagonal block of the Fisher has rank 600,000, while the approximation has rank 1. In convolutional networks this is even more pronounced: The full rank of the Fisher is multiplied by the number of locations at which the filer is applied, while the approximation remains at rank 1.

So in general, this approximation does not hold. In the literature it is usually justified by an independence assumption. Concretely, we view $\a_i\a_i^T, 
\e_i\e_i^T$ as random variables, where the randomness jointly depends on which datapoint we draw. We then assume that these random variables are independent. However, note that in non-degenerate neural networks we will usually be able to uniquely identify which datapoint was used, if we are given $\e_i\e_i^T$ (It is extremly unlikely that a back-propagated derivative is the same for two different datapoints -- so there is a one-to-one mapping from datapoints to $\e_i$). Thus we can uniquely determine $\a_i\a_i^T$. This implies that the conditional entropy $H(\a_i\a_i^T \mid \e_i\e_i^T) = 0$, in particular $\a_i\a_i^T, \e_i\e_i^T$ are not independent.

We point out that none of our experiments directly checks how accurate this approximation of the Fisher Information is. Our experiments mainly show that heuristic damping breaks the link to the Fisher, but do not give data on the quality of the approximation before heuristic damping. Evaluating the quality of this approximation is an interesting question left to future work. 

Relating this to findings of \cite{bernacchia2019exact}, note that if we consider linear networks and regression problems with homoscedastic noise, then the backpropaged derivatives $\e_i$ are completely independent of the datapoint -- and in particular independent of $\a_i$. This is because the derivatives at the last layer are a function of the covariance matrix of the noise (and independent of the in-/output of the network), and all derivatives are backpropagated through the same linear network. This is a part of the insights from \cite{bernacchia2019exact} for the analysis of linear networks. 
The above reasoning also explains precisely where this breaks down for non-linear networks (or hetero-scedastic noise). 
In particular, in non-linear networks, the errors will be backpropagated through different non-linear functions (given by the activations from the forward pass) and the argument from the linear case breaks down. 

\section{Toy Example Illustrating the Difference between SGD and FOOF}
\label{sec:toy_foof}
In the main paper, we pointed out that FOOF trades-off conflicting gradients differently (seemingly better) than SGD. Here, we provide a toy example illustrating this point. 
Roughly, the example will show that in SGD, gradients from one datapoint can "overwrite" gradients from other datapoints, so that SGD does not decrease the loss on the latter, while in FOOF the update will make progress on both datapoints simultaneously. 

The example will consist of a linear regression network with two inputs and one output and will feature two datapoints. The datapoints have inputs (3,1) and (1,0) and labels (1), (-1) and the weight vector (consisting of two weights) is initialised to (0,0). Taking the squared distance as loss function, It is easily verified that the gradients for the datapoitns are (3,1) and (-1,0). The SGD update direction is (2, 1) and will increase the loss on the second datapoint independent of step size. In contrast, the (un-damped) FOOF update is given by (-1, 4) which decreases the loss on both datapoints for suitably small stepsize. In particular a single update-step of FOOF with stepsize 1 converges to a global optimum.

In this context, it may be worth noting that applying FOOF to single datapoints and averaging resulting updates (rather than computing the FOOF update jointly on the  entire batch), corresponds to a version of SGD, in which each datapoint has a slightly different learning rate. We tested this version and found that it does not perform notably better than SGD. This also supports the intuition that FOOFs advantage over SGD comes from combining conflicting gradient directions more effectively.

\section{Kronecker-Factored Curvature Approximations for Laplace Posteriors}
\label{sec:kfac_laplace}
A Kronecker factored approximation of the curvature has also been used in the context of Laplace posteriors \cite{ritter2018scalable} and this has been applied to continual learning \cite{ritter2018online}. In both context, empirical results are very encouraging. 

Our finding that, in the context of optimization, the effectiveness of KFAC does not rely on its similarity to the Fisher raises the question whether these other applications \cite{ritter2018scalable, ritter2018online} of Kronecker-factorisations of the curvature rely on proximity to the curvature matrix.

We point out that the applications from \cite{ritter2018scalable, ritter2018online} do not seem to rely on heuristic damping. Also in light of our findings, it remains plausible that without heuristic damping, KFAC is very similar to the Fisher. In other words, the below is a hypothesis, not a certainty. To evaluate it, it would be interesting to compare the performances of KFAC to Full Laplace as well as to the algorithm suggested below. 

For simplicity of notation, we assume that the network has only one layer, but the analysis straightforwardly generalises to more layers.
Suppose $\W_0$ is a local minimum of the negative log-likelihood of the parameters. Further denote the approximation to the posterior covariance by $\Sig$. 
For an approximate posterior to be effective, we require that parameters which are assigned high likelihood by the posterior actually do have high likelihood according to the data distribution.
In other words, when a weight pertubation $\V$ satisfies that $\vec{\V}^T\Sig \vec{\V}$ is small (high likelihood according according to the approximate posterior), then the parameter $\W_0+\V$ should have low loss (i.e.\ high likelihood according to the data). 

For the Kronecker-factorisation, the first factor again is given by $\A\A^T$. Let us assume again that the second factor is dominated by a damping term (a very similar argument works if the second factor is predominantly diagonal), so that the posterior covariance is approximately $\Sig \approx \A\A^T\otimes \I$. Then, some easily verified calculations give 

\begin{align}
	\vec{\V}^T \Sig \vec{\V} \approx \vec{\V}^T (\A\A^T\otimes\I)\vec{\V} = \sum_i \v_i^T (\A\A^T) \v_i
\end{align}
where $\v_i$ is the $i$-th row of $\V$, i.e. the set of weights connected to the $i$-th output neuron.\footnote{If the second kronecker-factor is not the identity, then there are additional cross terms of the form $b_{ij}\v_i^T (\A\A^T) \v_j$, where $b_{ij}$ is the $i,j$-th entry of the second kronecker-factor.}
This expression being small means that each row of the pertubation $\V$ is near orthogonal to the input activations $\A$ (or more formally, it aligns with singular vectors of $\A$ which correspond to small singular values). This means that the layer's output, and consequently the networks output, get perturbed very little. This in turn means, that $\W_0+\V$ has high-likelihood.

A simple test of this hypothesis would be to keep only the first kronecker-factor $\A\A^T$, replace the second one by the identity and check if the method performs equally well or better. Further, it would be interesting to compare the performance of kronecker-factored posterior to a full-laplace posterior (controlling for the amount of data given to both) and check if -- analogous to our results for optimization -- the kronecker-factored posterior outperforms the exact laplace posterior. 

In fact, a very similar algorithm has already been developed independently \cite{ober2021global}. It shows strong performance, indirectly supporting our hypothesis.

\section{Related Work}\label{sec:related}
We review generally related work as well as more specifically algorithms with similar updates rules to FOOF.

\subsection{Generally Related Work}

Natural gradients were proposed by Amari and colleagues, see e.g.\ \cite{amari1998natural} and its original motivation stems from information geometry \cite{amari2000methods}. It is closely linked to classical second-order optimization through the link of the Fisher to the Hessian and the Generalised Gauss Newton matrix \cite{martens2014new, pascanu2013revisiting}.  Moreover, natural gradients can be seen as a special case of Kalman filtering \cite{ollivier2018online}. Interestingly, different filtering equations can be used to justify Adam's \cite{kingma2014adam} update rule \cite{aitchison2018bayesian}, see also \cite{khan2018fast}.

There is a long history of approximating natural gradients and second order methods.
For example, HF \cite{martens2010deep} exploits that Hessian-vector products are efficiently computable and uses the conjugate-gradient method to approximate products of the inverse Hessian and vectors. In this case, similarly to our application, the Hessian is usually subsampled, i.e.\ evaluated on a mini-batch.
Other approximations of natural gradients include \cite{roux2007topmoumoute, ollivier2015riemannian, ollivier2017true, grosse2015scaling, desjardins2015natural, martens2010deep, marceau2016practical}. 

The intrinsic low rank structure of the (empirical) Fisher has been exploited in a number of setups by a number of papers including \cite{agarwal2019efficient, goldfarb2020practical, immer2021scalable, dangel2021vivit}. 

Kronecker-factored approximations \cite{martens2015optimizing, grosse2015scaling} have become the basis of several optimization algorithms \cite{botev2017practical, goldfarb2020practical, george2018fast, bernacchia2019exact}. Our contribution may shed light on why this is the case.

Moreover, Kronecker-factored approximation of the curvature can be used in the context of Laplace Posteriors \cite{ritter2018scalable}, which can also be applied to continual learning \cite{ritter2018online}. A more detailed discussion of how this relates to our findings can be found in Section \ref{sec:kfac_laplace}.

KFAC faces the problem of approximating a sum of kronecker-products by a single kronecker-product. This problem also occurs when approximating real time recurrent learning of recurrent networks \cite{williams1995gradient, tallec2017unbiased, mujika2018approximating, benzing2019optimal} and in this context \cite{benzing2019optimal} show how to obtain optimal biased and unbiased approximations. Our results suggest that it is not promising to apply these techniques to approximate natural gradients more accurately.

As briefly mentioned, FOOF is related to the idea of optimizing modules of a nested function independently, e.g.\ \cite{lecun1988theoretical, carreira2014distributed, taylor2016training, gotmare2018decoupling}.

It may also be worth noting that FOOF is evocative of target propagation \cite{bengio2014auto, meulemans2020theoretical}, but we are not aware of a formal link between these methods.

\subsubsection{Comparison between HF and KFAC}
\label{sec:hf_vs_kfac}
The subsampled natural gradient method upon which many of our results rely was first described in \cite{ren2019efficient}. On top of their useful, important theoretical results, they also provide an empirical evaluation of their method, and -- to the best of our knowledge -- the only published comparison of KFAC and HF. 

Unfortunately, there is very strong evidence that all methods considered there are heavily undertuned or that there is another issue. To see this, note that in \cite{ren2019efficient} a network trained on MNIST with one hidden layer of 500 neurons achieves a training loss of around 0.3 and a test accuracy of less than 95\% for all considered optimizers. 
This is much worse than standard results and clearly not representative of normal neural network training. We quickly verified that in exactly the same setting, with KFAC we are able to obtain a loss which is more than 100x smaller and a test accuracy of 98\%.\footnote{
	We used the same network, same activation function and same number of epochs. Weight initialisation scheme, batch size and preprocessing were not described in the original experiments, so we used batch size 100 and the same initialisation and preprocessing as in our other MNIST experiments (which has no data augmentation).}

\subsubsection{Theoretical Work}
There also is a large body of work on theoretical convergence properties of Natural Gradients. 
We give a brief, incomplete overview here and 
refer to \cite{zhang2019fast} for a more thorough discussion.

\cite{bernacchia2019exact} analyse the convergence of natural gradients in linear networks. Interestingly, they show that for linear networks applied to regression problems (with homoscedastic noise), inverting a block-diagonal, Kronecker-Factored approximation of the curvature results in exact natural gradients, see also Appendix~\ref{sec:kfac_derivation} for a brief justification of a part of their findings. We point out that the empirical results in non-linear networks from \cite{bernacchia2019exact} essentially amount to a re-discovery of KFAC, as such they do not contradict our results. In particular, they are also based on heuristic damping.

For non-linear, strongly overparametrised two-layer networks in which only the first layer is trained, \cite{zhang2019fast} recently gave a convergence analysis of both natural gradients and KFAC. 
Note that \cite{zhang2019fast} do not establish similarity between KFAC and Natural Gradients but rather give two separate convergence proofs. 

Both these theoretical results \cite{bernacchia2019exact, zhang2018noisy} do not account for any form of damping, so they have to be seen as independent of the empirically well-performing version of KFAC and our investigation.


A set of interesting theoretical results by \cite{karakida2021pathological} shows that the Fisher Information in deep neural networks has a pathological spectrum -- in particular, they show that the Fisher is flat in most directions. This view may well give a theoretical intuition for why Subsampled Natural Gradients do often not notably outperform SGD. 

\subsubsection{Related Work from Bayesian ML}
Similar to our new view on optimization is \cite{ober2021global}, which is a Bayesian Posterior approximation and can (roughly) be viewed as considering distributions over neuron activations rather than in weight space directly, similarly to how FOOF performs optimization steps on neuron activations rather than on weights directly. 

\subsection{Algorithms with similar update rules}\label{sec:related_similar}

While it is not immediately visible due to a re-parametrisation employed in \cite{desjardins2015natural}, Natural Neural Networks \cite{desjardins2015natural} (NNN)
propose a mathematically very similar update rule to FOOF (and KFAC). Unlike FOOF, NNN centers layer inputs by subtracting the mean activation (or an estimate thereof), but like FOOF they ignore the second kronecker factor of KFAC. 

Like KFAC, NNN is derived as a block-diagonal, kronecker-factored approximation of the Fisher. As we already pointed out, this is very puzzling, since NNN approximates the Fisher by a zero-th order matrix, ignoring all first- and second-order information. In this sense, and bearing in mind our previous experiments, NNN should not be seen as a natural gradient method and our results offer an explanation why it is nevertheless so effective. 

From an implementational viewpoint, FOOF is preferable to NNN mainly because it requires inverting matrices rather than computing SVDs. In practice computing inverses is both considerably faster (a factor of 10 or so as found in some quick experiments) and more stable  then computing the SVD, as NNN does (SVD algorithms don't always converge).

With yet another context and motivation, \cite{frerix2017proximal} also proposes a similar update rule focussing on full-batch descent. The motivation can be roughly rephrased as imposing proximity constraints on neuron activations. Very recently, their motivation and algorithm seems to have been re-described in \cite{amid2021locoprop} without noting this link. 
In particular, the derivation of \cite{frerix2017proximal} gives an alternative perspective on the update equation of FOOF. 

Among other differences, \cite{frerix2017proximal, amid2021locoprop} (1) seem not to discuss unbiased stochastic versions of their algorithms, (2) seem less computationally efficient: results in \cite{frerix2017proximal} fall short of adam in terms of wall-clock time and \cite{amid2021locoprop} does not provide direct wall-clock time comparisons with standard first-order optimizers, (3) only discuss fully-connected architectures (4) do not perform investigations into the connection of KFAC to natural gradients or first-order methods.

Note also that the framework from \cite{ollivier2018online} can be applied to interpret FOOF as applying Kalman filtering to each layer individually. Thus, in some vague sense, FOOF is bayes-optimal and some may find this to be an enticing explanation for FOOF's strong empirical performance.

\section{Details for Efficiently Computing $\F^{-1}$-vector products for a Subsampled Fisher}
\label{sec:nat_details}
\subsection{Notation}
For simplicity, we restrict the exposition here to fully connected neural networks without biases and to classification problems. Our method is also applicable to regression problems, can easily be extended to include biases and to handle for example convolutional layers.

We denote the network's weight matrices by $\W=\left(\W^{0}, \ldots, \W^{(\ell-1)}\right)$, where $W^{(i)}$ has dimensions $n_i \times n_{i+1}$. 
The pointwise non-linearity will be denoted $\sigma(\cdot)$. 
For a single input $\x=\a^{(0)}$ the network iteratively computes
\begin{align}
	s^{(k)} &= \W^{(k-1)}\a^{(k-1)}&&\text{for }k=1,\ldots,\ell  \\
	\a^{(k)} &= \sigma(\s^{(k)}) &&\text{for } k=1,\ldots,\ell-1 \\
	f(\x) = f(\x;\W)&=\mathrm{softmax}(a^{(\ell)})&&
\end{align}
We use the cross-entropy loss $L(f(\x), y)$ throughout. 
We will write $\e^{k} = \frac{\partial L(f(x, y))}{\partial s_k}$ for the errors, which are usually computed by backpropagation.

If we process a batch of data, we will use upper case letters for activations and ``errors'', i.e.\ $\A^{(k)}, \S^{(k)}, \E^{(k)}$ which have dimensions $n_k\times B$, where $B$ is the batch size. The $i$-th column of these matrices will be denoted by corresponding lower case letters, e.g.\ $\a^{(k)}_i$.

We will write $\A\odot \B$ for the pointwise (or Hadamard) product of $\A,\B$ and $\A\otimes\B$ for the Kronecker product. The euclidean inner product (or dot product) will be denoted by $\A\cdot\B$ for both vectors and matrices.

The number of parameters will be called $n=\sum_{k=0}^{\ell-1}{n_k n_{k+1}}$, the batch size $B$, the output dimension of the network $n_\ell$.

We will generally assume derivatives to be one dimensional column vectors and will often write $\g = \g(\x,y) = \frac{\partial L(f(\x),y)}{\partial \W}$ and $\g^{(k)} = \frac{\partial L(f(\x),y)}{\partial \W}$. Generally, for a vector $\u$ of dimension $n$, the superscript $\u^{(k)}$ will denote the entries of $\u$ corresponding to layer $k$, and $\mathrm{mat}\left(\u^{(k)}\right)$ will be a matrix with the same entries as $\u^{(k)}$ and of the same dimensions as $\W^{(k)}$.

\subsection{Overview}
The technique described here is similar to \cite{agarwal2019efficient, ren2019efficient}. However, the implementation of \cite{ren2019efficient} requires several for- and backward passes for each mini-batch, which is used to compute the Fisher, while \cite{agarwal2019efficient} uses the same ideas, but applies them in  a different context, which does not require computing the Fisher.
Another key difference, both in terms of computation time and update-direction quality (or bias of updates) is discussed in Section~\ref{sec:nat_bias}.

We now outline how to efficiently compute exact natural gradients under the assumption that the Fisher is estimated from a mini-batch of moderate size (where `moderate' can be on the order of thousands without large difficulties). Let's assume we have $B$ samples $(\x_1, y_1), \ldots, (\x_B, y_B)$ from the model distribution and use this for a MC estimate of the Fisher, i.e.
\begin{align}
	\F = \frac{1}{B}\sum_{i=1}^{B} \g_i \g_i^T = \G \G^T
\end{align}
where we define $\G$ to be the matrix whose $i$-th column is given by $\frac{1}{\sqrt{B}}\g_i$. An application of the matrix inversion lemma now gives
\begin{align}
	(\lambda \I+\F)^{-1}\u = (\lambda\I+\G \G^T)^{-1} = \lambda^{-1}\I\u - \lambda^{-2}\G (\I + \frac{1}{\lambda} \G^T \G)^{-1}\G^T\u 
\end{align}

We will not compute $\G$ explicitly. Rather, we will see that all needed quantities can be computed efficiently from the quantities obtained during a single standard for- and backward pass on the batch $\{(\x_i, y_i)\}_{i=1}^B$, namely the preactivations $\A^{(k)}$ and error $\E^{(k+1)}$.\footnote{We note that many of the required quantities can be seen as Jacobian-vector products and could be computed with autograd and additional for- and backward passes. Here, we simply store preactivations and errors from a single for- and backward pass to avoid additional passes through the model.}

Overall, the computation can be split into three steps. (1)~We need to compute $\v=\G^T \u$. (2)~We need to compute $\G^T \G$, after which evaluating $\w = (\I + \frac{1}{\lambda}\G^T\G)^{-1}\v$ is easy by explicitly computing the inverse. (3)~We need to evaluate $\G \w$.

Very briefly, the techniques to compute (1)-(3) all rely on the fact that, for a single datapoint, the gradient with respect to a weight matrix is a rank 1 matrix and that, consequently, gradient-vector and gradient-gradient dot products can be computed and vectorised efficiently. 

\subsection{Details}
We now  go through the steps (1)-(3) described above. 

For (1), note that the $i$-th entry of $\v = \G^T\u$ is the dot-product between $\g_i$ and $\u$, which in turn is the sum over layer-wise dot-products $\g_i^{(k)} \cdot \u^{(k)}$.
Note that $\mathrm{mat}\left(\g_i^{(k)}\right) = \a^{(k)}_i \e_i^{(k)T}$ is a rank one matrix, so that $\g_i^{(k)}\cdot\u^{(k)} = \a^{(k)T}_i \mathrm{mat}\left(\u^{(k)}\right)\e^{(k)}$. A sufficiently efficient way to vectorise these computations is the following:
\begin{align}
	\label{eq1}
	\v = \G^T \u = \sum_{k=0}^{\ell-1} \mathrm{diag}\left(\A^{(k),T} \mathrm{mat}\left(\u^{(k)}\right) \E^{(k+1)}\right)
\end{align}

For (2), similar considerations give
\begin{align}
	\label{eq2}
	\G^T\G  = \sum_{k=0}^{\ell-1} (\A^{(k)T} \A^{(k)}) \odot (\E^{(k+1)T} \E^{(k+1)})
\end{align}

Finally, for (3), note that $\G \w$ is a linear combination of the gradients (columns) of $\G$. Writing $\mathbf{1}$ for a column vector with $n_k$ ones, this can be computed layer-wise as
\begin{align}
	\label{eq3}
	\mat\left((\G\w)^{(k)}\right) = \A^{(k)T} \left(\E \odot (\w \mathbf{1}^T)\right)
\end{align}

We re-emphasise that we only require to know $\A^{(k)}$ and $\E^{(k+1)}$, which can be computed in a single for- and backward pass and then stored and re-used for computations. 

\subsection{Computational Complexity}
In terms of memory, we need to store $\A^{k}, \E^{k}$, which requires at most as much space as a single backward pass. Storing $\G^T\G$ requires space $B\times B$, which is typically negligible. as are results of intermediate computation. 

In terms of time, computations \eqref{eq1} takes time $O(B\sum_k n_k^2))$, \eqref{eq2} takes time $O(B \sum_k n_k^2)$, \eqref{eq3} requires $O(Bn)$. The matrix inverision requires $O(B^3)$, but note that technically we only need to evaluate the product of the inverse with a single vector, which theoretically can be done slightly faster (so can some of the matrix multiplications).


\subsection{Less biased Subsampled Natural Gradients}\label{sec:nat_bias}
Our aim is to estimate $(\lambda\I +\F)^{-1}\g$ and we use mini-batches estimates $\bar{\F}$ and $\bar{\g}$. 
Ideally, we would want an unbiased estimate, i.e.\ an estimate with mean $(\lambda\I+\EE[ \bar{\F}])^{-1}\cdot\EE[\bar{\g}]$.

One problem that seems hard to circumvent is that, while our estimate $\bar{\F}$ of $\F$ is unbiased, the expectation of $(\lambda \I + \bar{\F})^{-1}$ will not be equal to $(\lambda\I+\F)^{-1}$. We shall not resolve this problem here and simply hope that its impact is not detrimental.

Another problem is that using the same mini-batch to estimate Fisher $\F$ and gradient $\g$ will introduce additional bias: 
Even if $X = (\lambda\I+\bar{\F})^{-1}$ were an unbiased estimate of $(\lambda\I+\F)^{-1}$ and $Y = \bar{\g}$ is an unbiased estimate of $\g$, it does not automatically hold that $XY$ is an unbiased estimate of $\EE[X]\EE[Y]$. This does however hold, if $X,Y$ are independent, which can be achieved by estimating them based on independent mini-batches.
The fact that a bias of this kind can meaningfully affect results, also in the context of modern neural networks and standard benchmarks, has already been observed in \cite{benzing2020unifying}.

Thus, we propose using independent mini-batches to estimate $\bar{\F}$ and $\bar{\g}$. On top of removing bias from our estimate, this has the additional benefit that we do not have to update $\bar{\F}$ (or rather the quantities related to it) at every time step. This gives further computational savings. 

\begin{figure*}[h!]
	\centering
	\includegraphics[width=0.85\textwidth]{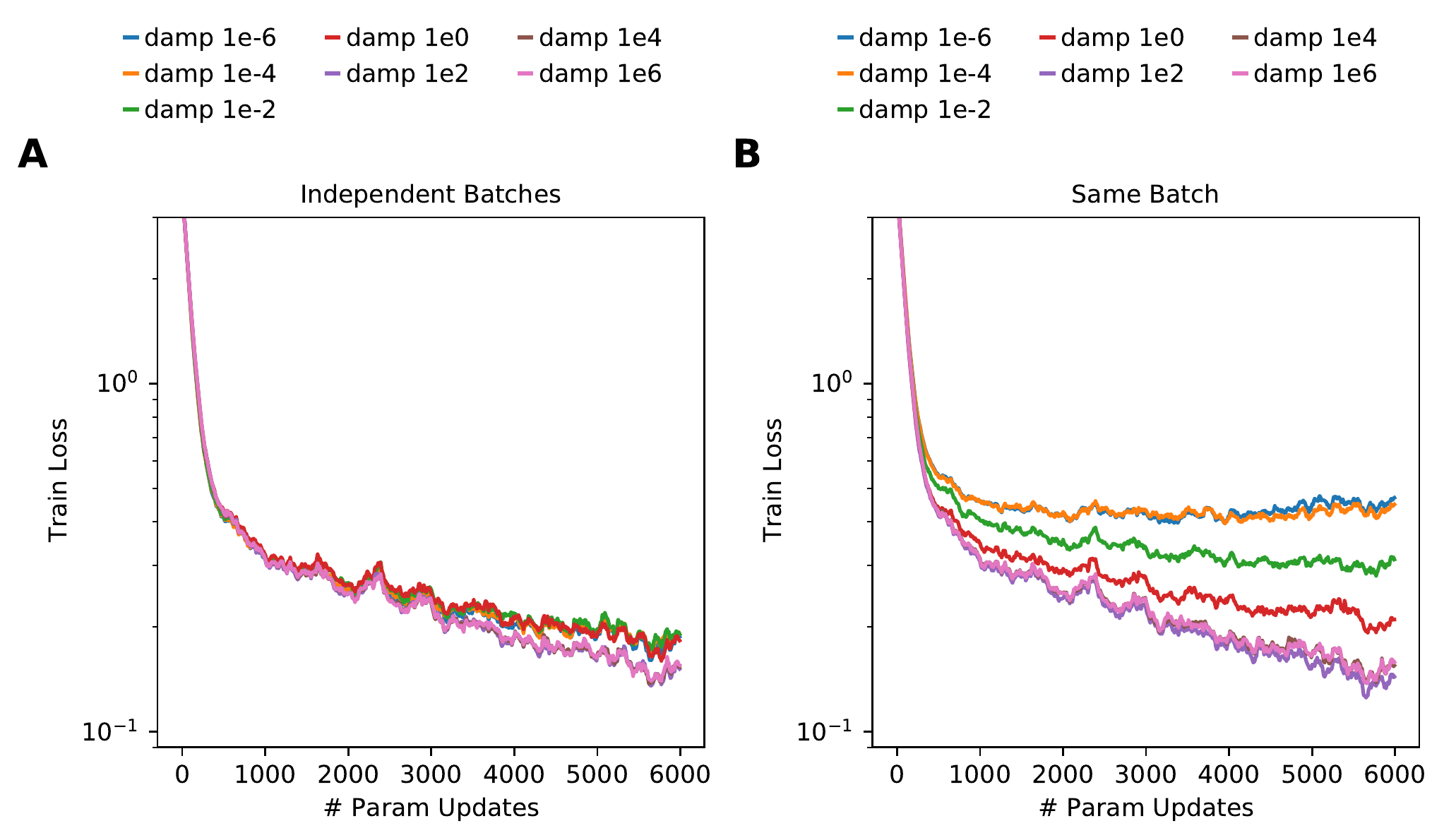}
	\vskip -15pt
	\caption{Investigating the effect of evaluating Fisher and gradient on different respectively identical minibatches.
	}
	\label{fig:nat_ab_f}
\end{figure*}

\begin{figure*}[h!]
	\centering
	\includegraphics[width=0.85\textwidth]{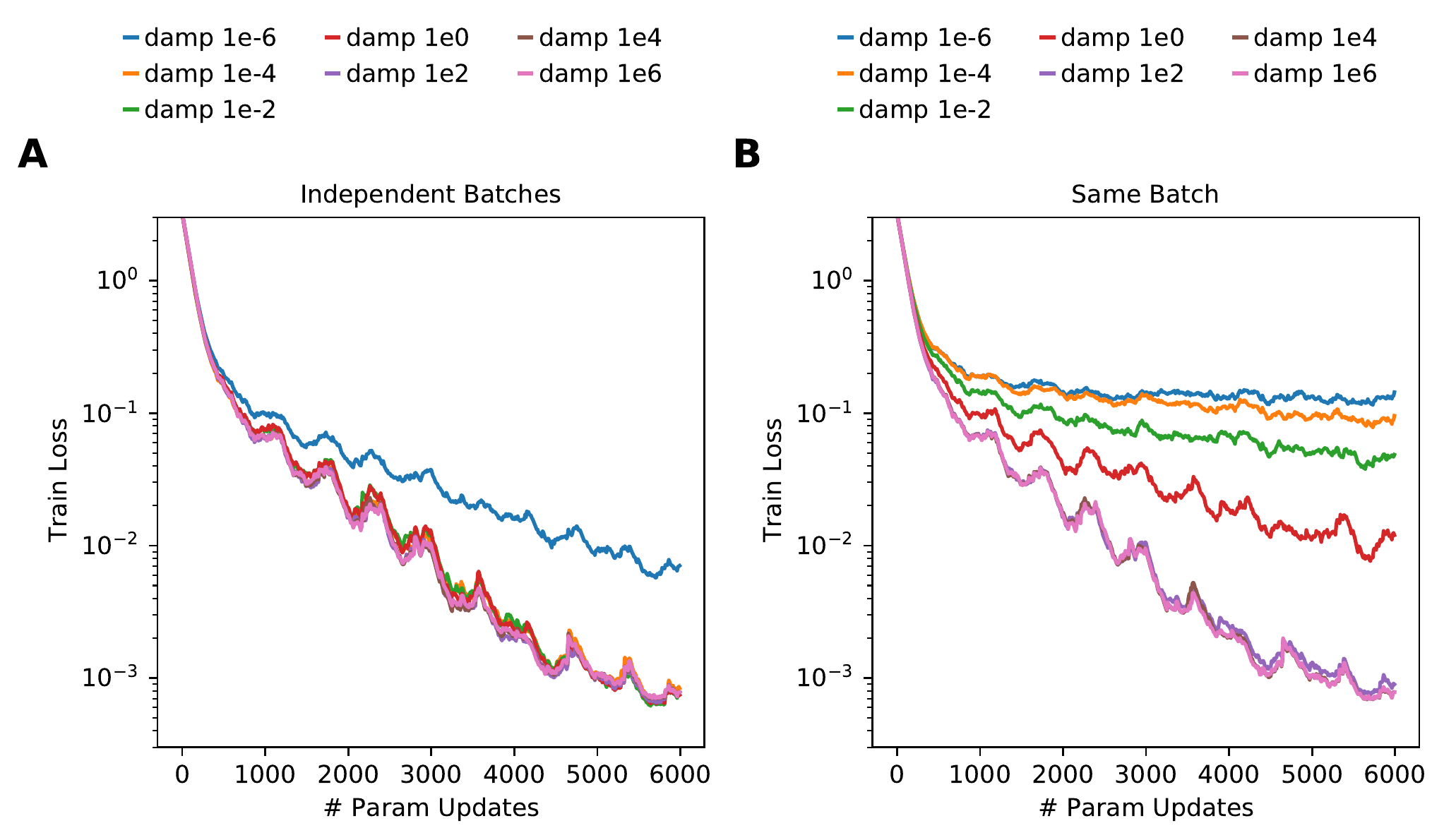}
	\vskip -15pt
	\caption{Same as Figure \ref{fig:nat_ab_f}, but on MNIST.
	}
	\label{fig:nat_ab_m}
\end{figure*}

We perform an ablation experiment for this choice in Figures~\ref{fig:nat_ab_f}~and~\ref{fig:nat_ab_m}. 


\clearpage
\section{Additional Experiments}\label{sec:additional_experiments}
Here we present additional experiments, lines correspond to the average across three seeds.

In Section~\ref{sec:additional_mnist}, there are experiments analogous to Figures~\ref{fig:kfac_ablation_data}-\ref{fig:foof} from the main paper, but on MNIST rather than Fashion-MNIST. Results are in line with the ones on Fashion-MNIST, but effects are usually smaller, presumably due to MNIST-classification being a very simple task for MLPs. 

In Section~\ref{sec:additional_subsampled_kfac}, there are comparisons of subsampled natural gradients to subsampled KFAC for a ResNet on CIFAR10 and for autoencoder experiments. The fact that KFAC performs considerably better than natural gradients strongly suggests that also in these settings KFAC cannot be seen as a natural gradient method. 

In Section~\ref{sec:additional_damping}, we show that heuristic damping is crucial for performance of KFAC for a ResNet on CIFAR10 and for autoencoder experiments. This further supports the claim that KFAC should not be seen as a natural gradient method and suggests that similarity to FOOF is important for KFAC.

In Section~\ref{sec:additional_performance}, we show performance comparisons between KFAC and FOOF for different benchmarks and architectures. 
Figure~\ref{fig:cifar100} contains performance of a Wide ResNet18 on CIFAR 100 (rather than CIFAR 10 in the main paper).
Figures~\ref{fig:vgg_1}, \ref{fig:vgg_2} contain training data for a VGG11 network on SVHN and additionally show how different algortihms are affected by different batch sizes. 

\subsection{Limitations of our Explanation in Auto-Encoder Settings}
Figures~\ref{fig:faces}-\ref{fig:curves} contain autoencoder experiments on MNIST, Curves and Faces. Here, we make a somewhat puzzling observation: When we follow the KFAC training trajectory, FOOF makes more progress per parameter udpate than KFAC. Nevertheless, when we use FOOF for training (and follow the FOOF trajectory), we obtain slightly worse results than for KFAC. This may suggest that in the autoencoder experiments KFAC chooses a different trajectory that is easier to optimize than FOOF.\\ 
This suggests that our explanation of KFACs performance, while capturing many key-characteristics of KFAC, has some limitations.
\\
We re-emphasise that similarity to FOOF remains a significantly better explanation for KFAC's performance than similarity to natural gradients, also in the autoencoder setting (recall Figure~\ref{fig:auto_subsampled}). \\
As an additional experiment we run KFAC with the empirical Fisher, rather than an MC approximation. Despite its name, the empirical Fisher is usually argued to be a poor approximation of the Fisher \cite{martens2014new, kunstner2019limitations}. Nevertheless we find that KFAC works equally well with the empirical Fisher, see Figure~\ref{fig:auto_kfac_emp}, supporting the view that KFAC's effectiveness is not directly linked to the Fisher and hinting at the fact that a full, principled explanation of KFAC's slight advantage over FOOF in the autoencoder setting may be difficult to come by.

\clearpage
\subsection{Experiments anaologous to main paper but on MNIST rather than Fashion MNIST}
\label{sec:additional_mnist}
\begin{figure*}[h]
	\centering
	\includegraphics[width=0.85\textwidth]{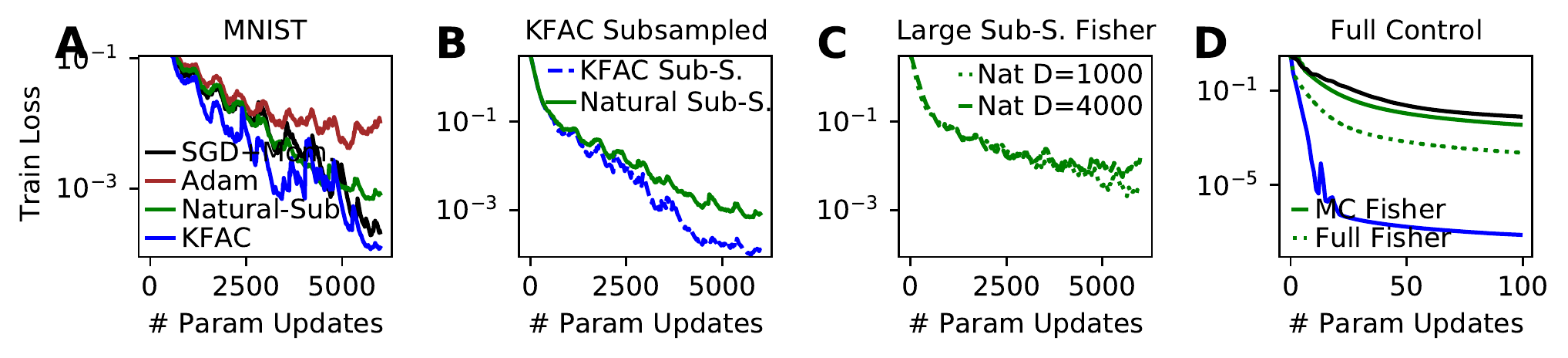}
	\vskip -10pt
	\caption{
		\textbf{Paradoxically, KFAC -- an approximate second-order method -- outperforms exact second-order udpates in standard as well as important control settings.}
		Same as Figure~\ref{fig:kfac_ablation_data} but on MNIST rather than Fashion MNIST.
	}
	\label{fig:kfac_ablation_data_mnist}
\end{figure*}

\begin{figure*}[h]
	\centering
	\includegraphics[width=0.85\textwidth]{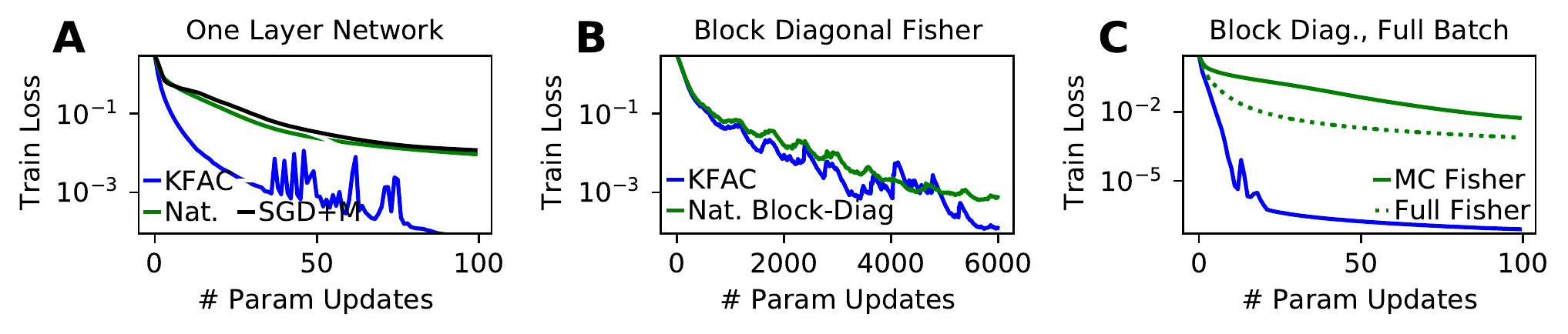}
	\vskip -10pt
	\caption{\textbf{Advantage of KFAC over exact, subsampled Natural Gradients is not due to block-diagonal structure.} 
		Same as Figure~\ref{fig:kfac_ablation_bd} but on MNIST rather than Fashion MNIST.
	}
	\label{fig:kfac_ablation_bd_mnist}
	\vskip -0pt
\end{figure*}

\begin{figure*}[h]
	\centering
	\includegraphics[width=.85\textwidth]{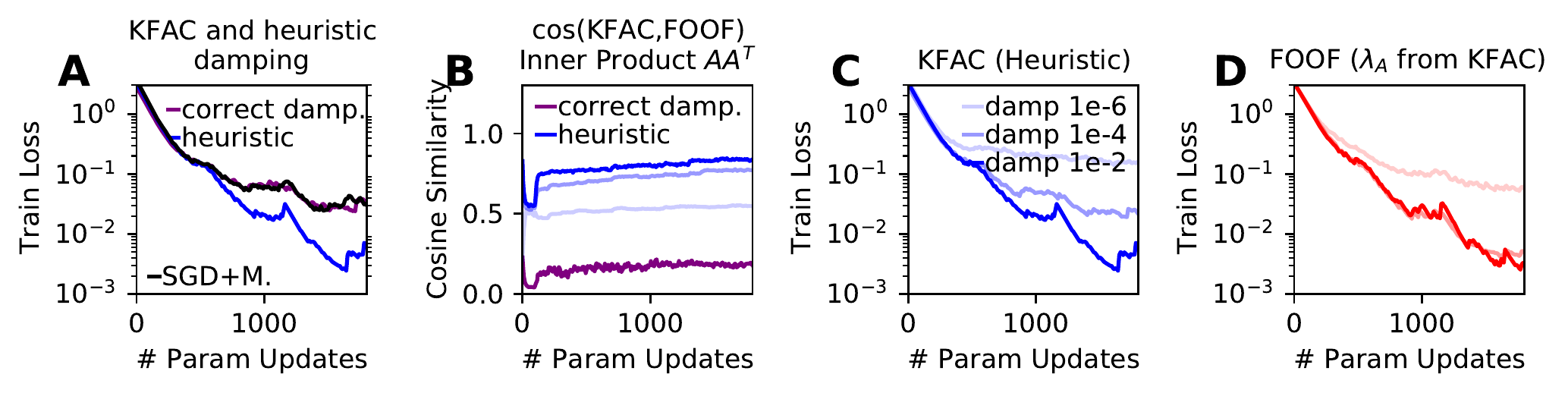}
	\vskip -10pt
	\caption{\textbf{Heuristic Damping increases KFAC's performance as well as its similarity to first-order method FOOF.}
		Same as Figure~\ref{fig:kfac_vs_first} but on MNIST rather than Fashion MNIST.
	}
	\label{fig:kfac_vs_first_mnist}
	\vskip -0pt
\end{figure*}

\begin{figure*}[h!]
	\vskip -0pt
	\centering
	\includegraphics[width=0.85\textwidth]{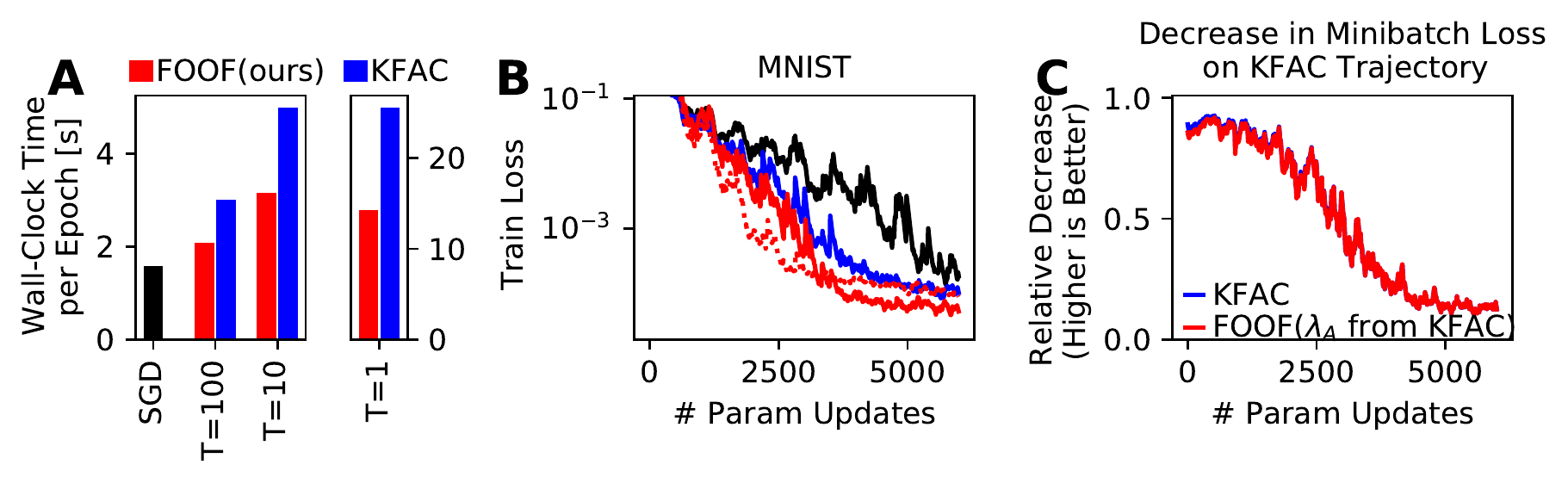}
	\vskip -10pt
	\caption{\textbf{FOOF outperforms KFAC in terms of both per-update progress and computation cost.} 
		Same as Figure~\ref{fig:foof} but on MNIST rather than Fashion MNIST.
	}
	\label{fig:foof_mnist}
	\vskip -0pt
\end{figure*}

\clearpage
\subsection{Subsampled Natural Gradients vs Subsampled KFAC}
\label{sec:additional_subsampled_kfac}

\begin{figure*}[h]
	\centering
	\includegraphics[width=0.95\textwidth]{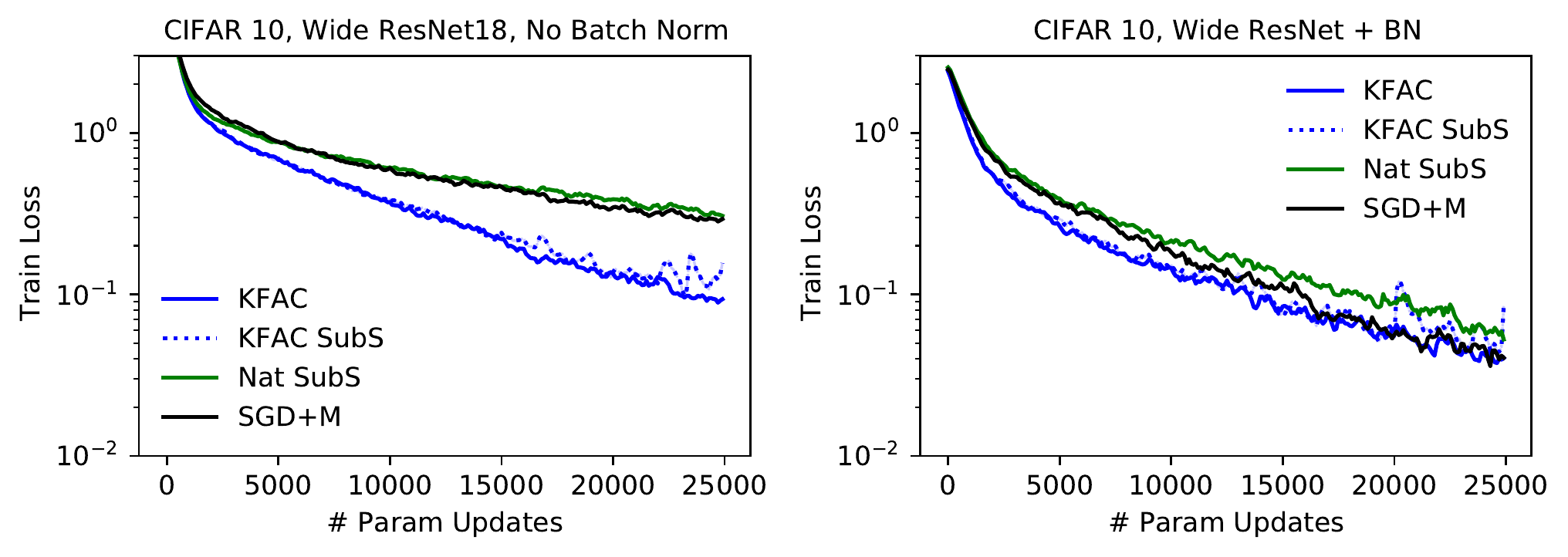}
	\vskip -15pt
	\caption{KFAC outperforms Subsampled Natural Gradients, also when KFAC is subsampled and uses exactly the same amount of data as Natural gradients to estimate the curvature. This  is analogous to Figure~\ref{fig:kfac_ablation_data}B, but on ConvNets and with a more complicated dataset. It confirms our claim that KFAC does not rely on second-order information. Note that with large damping, natural gradients becomes approximately equal to SGD -- thus the difference seen between SGD+M and natural gradients is due to momentum.
	}
	\label{fig:cifar_control}
\end{figure*}

\begin{figure*}[h]
	\centering
	\includegraphics[width=0.95\textwidth]{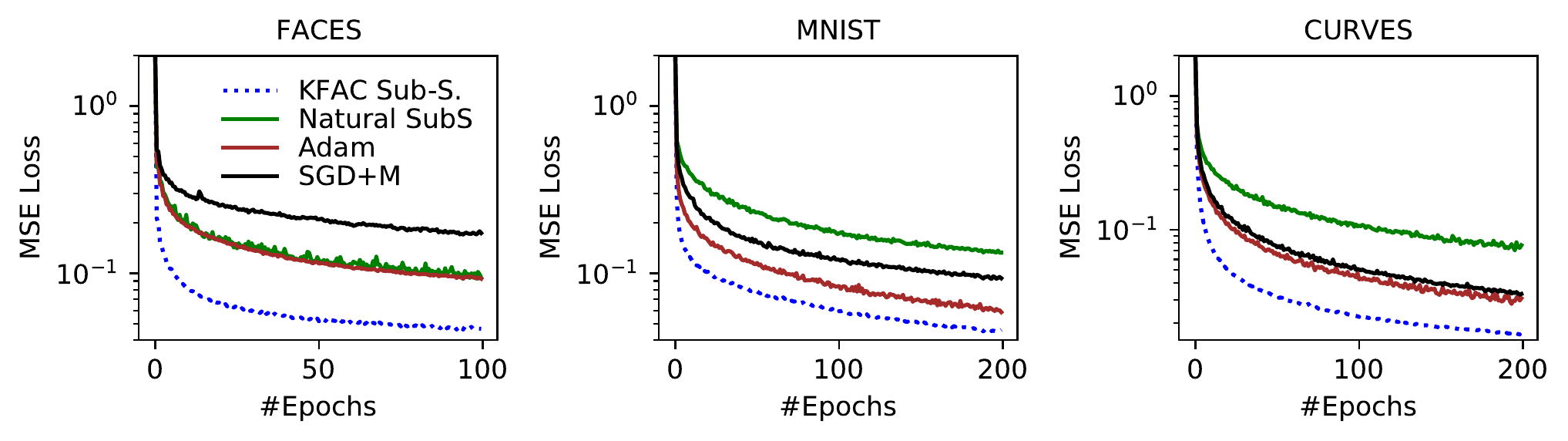}
	\vskip -15pt
	\caption{KFAC outperforms Subsampled Natural Gradients, also when KFAC is subsampled and uses exactly the same amount of data as Natural gradients to estimate the curvature. Autoencoder Experiments. We confirmed that Natural Gradients do not perform worse than SGD without momentum. In other words the advantage of SGD+M vs Natural Gradients on MNIST and Curves is due to using momentum.
	}
	\label{fig:auto_subsampled}
\end{figure*}

\clearpage
\subsection{Effect of Heuristic Damping on KFAC}
\label{sec:additional_damping}
\begin{figure*}[h]
	\centering
	\includegraphics[width=0.95\textwidth]{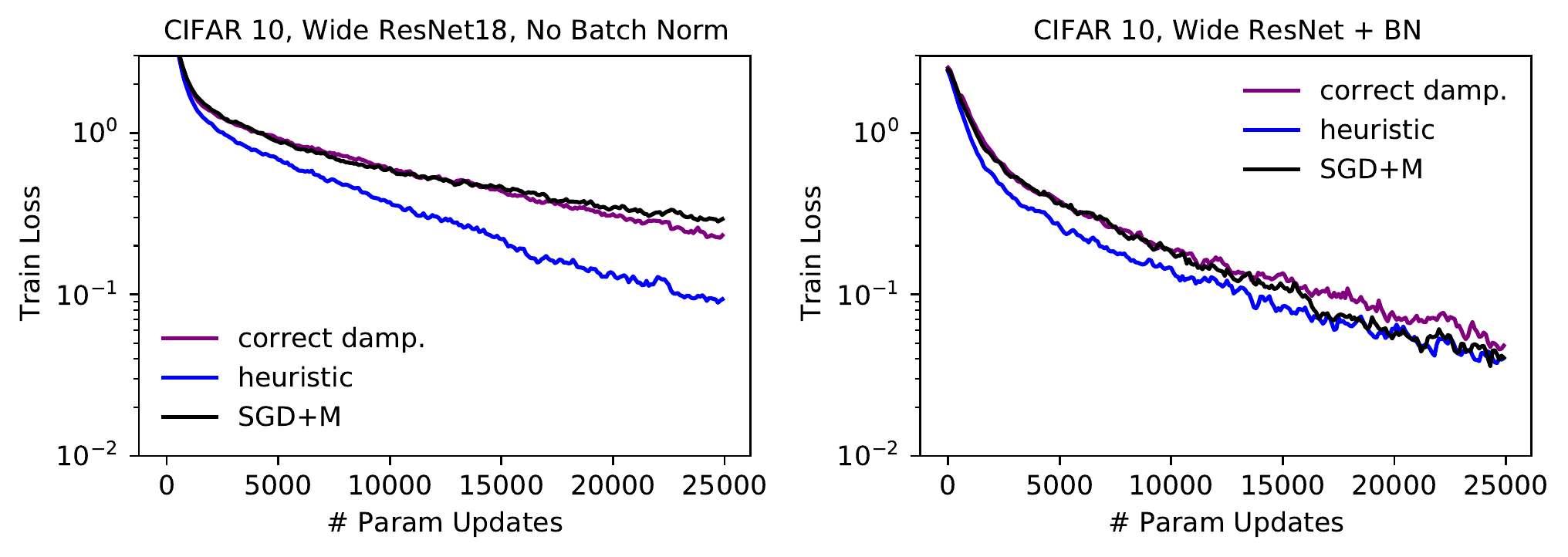}
	\vskip -15pt
	\caption{Effect of Heuristic damping on KFAC on CIFAR10 with a ResNet. Analogously to Figure~\ref{fig:kfac_vs_first}A, we find that heuristic damping is essential for KFAC's performance.
	}
	\label{fig:cifar_damp}
\end{figure*}

\begin{figure*}[h]
	\centering
	\includegraphics[width=0.95\textwidth]{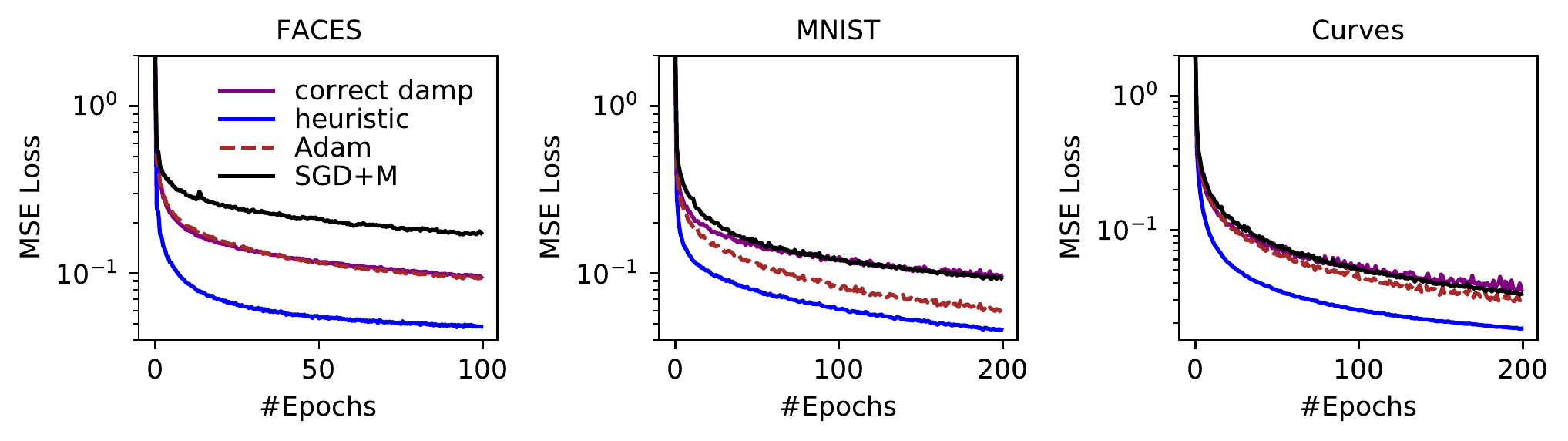}
	\vskip -15pt
	\caption{Effect of Heuristic damping on KFAC in autoencoder experiments. Analogously to Figure~\ref{fig:kfac_vs_first}A, we find that heuristic damping is essential for KFAC's performance.
	}
	\label{fig:auto_damp}
\end{figure*}

\clearpage
\subsection{Performance Comparisons}
\label{sec:additional_performance}
\begin{figure*}[h]
	\centering
	\includegraphics[width=0.95\textwidth]{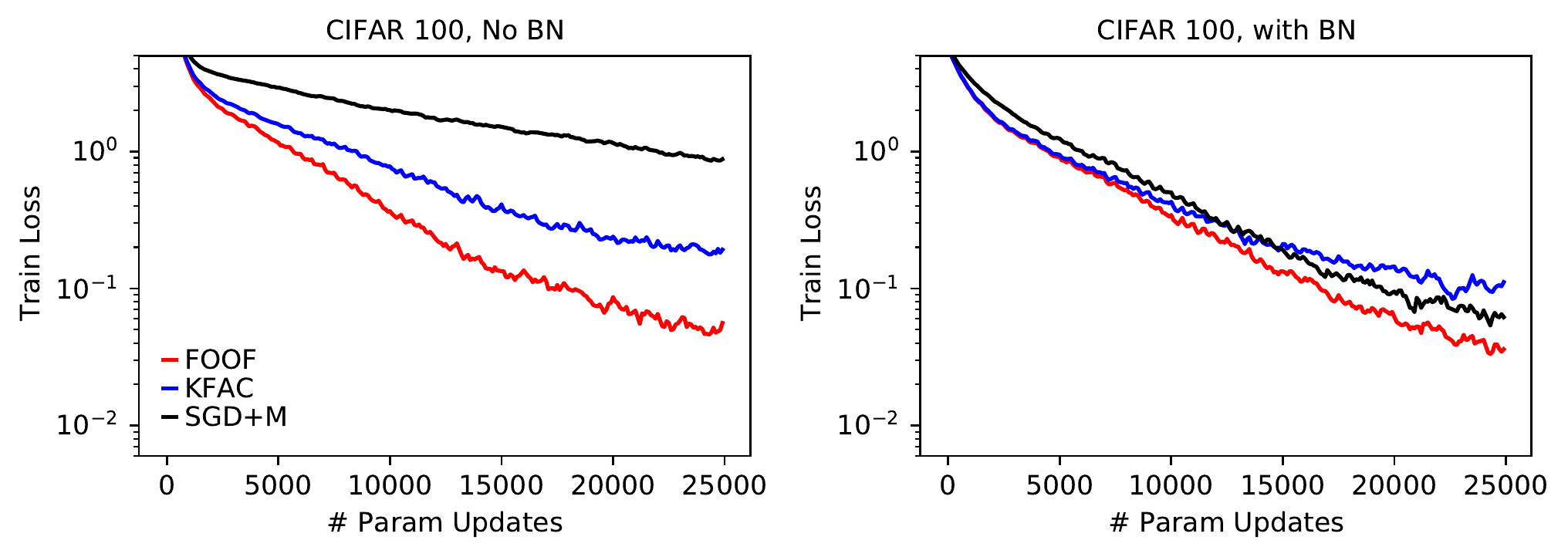}
	\vskip -15pt
	\caption{Performance comparison on CIFAR 100 with a Wide ResNet18.
	}
	\label{fig:cifar100}
\end{figure*}
\begin{figure*}[h]
	\centering
	\includegraphics[width=0.95\textwidth]{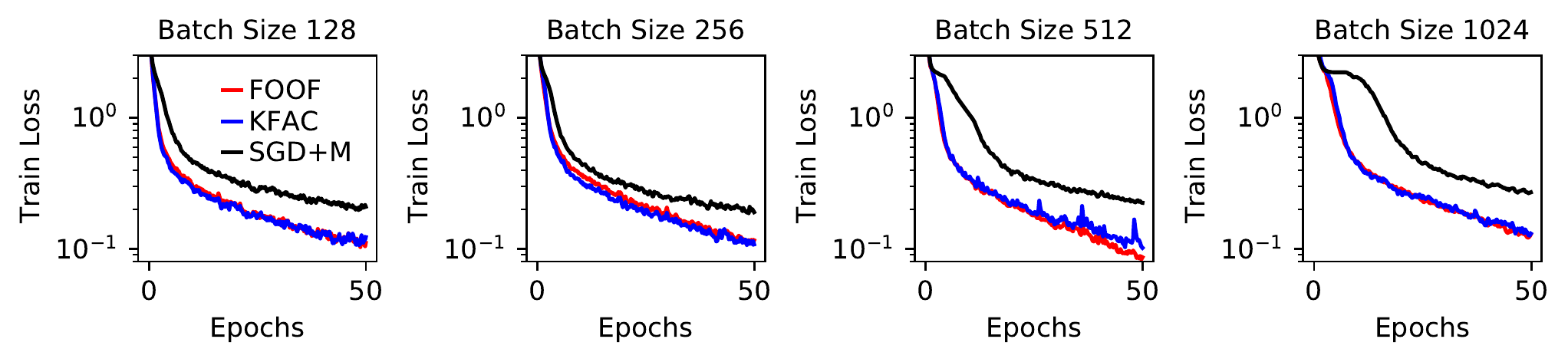}
	\vskip -15pt
	\caption{Performance comparison on SVHN with a VGG11 network and different batch sizes. See also Figure~\ref{fig:vgg_2} for same data portrayed differently.
	}
	\label{fig:vgg_1}
\end{figure*}
\begin{figure*}[h]
	\centering
	\includegraphics[width=0.95\textwidth]{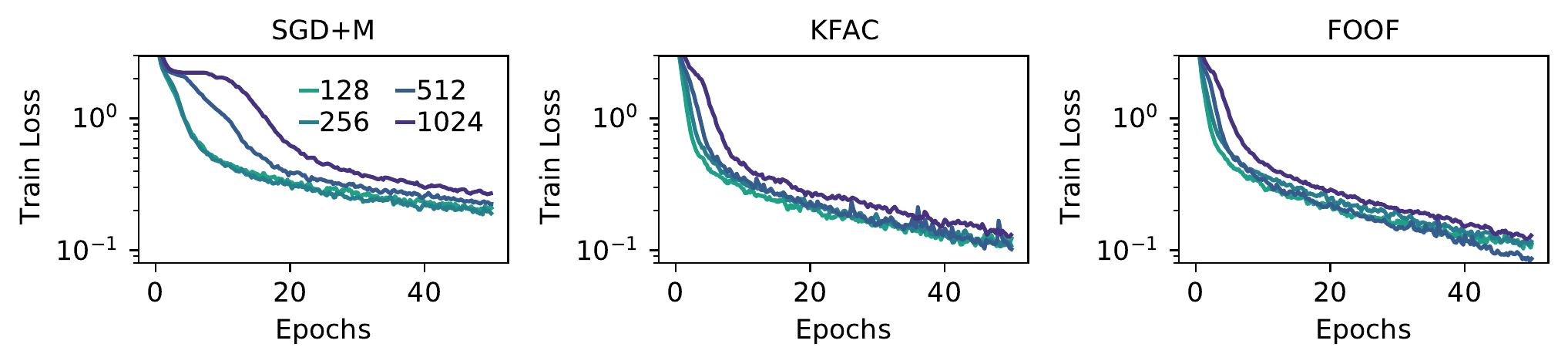}
	\vskip -15pt
	\caption{Performance on SVHN with a VGG11 network across different batch sizes for different algorithms. See also Figure~\ref{fig:vgg_1} for same data portrayed differently. Note that color coding differs from remaining plots in the paper.
	}
	\label{fig:vgg_2}
\end{figure*}

\begin{figure*}[h]
	\centering
	\includegraphics[width=0.95\textwidth]{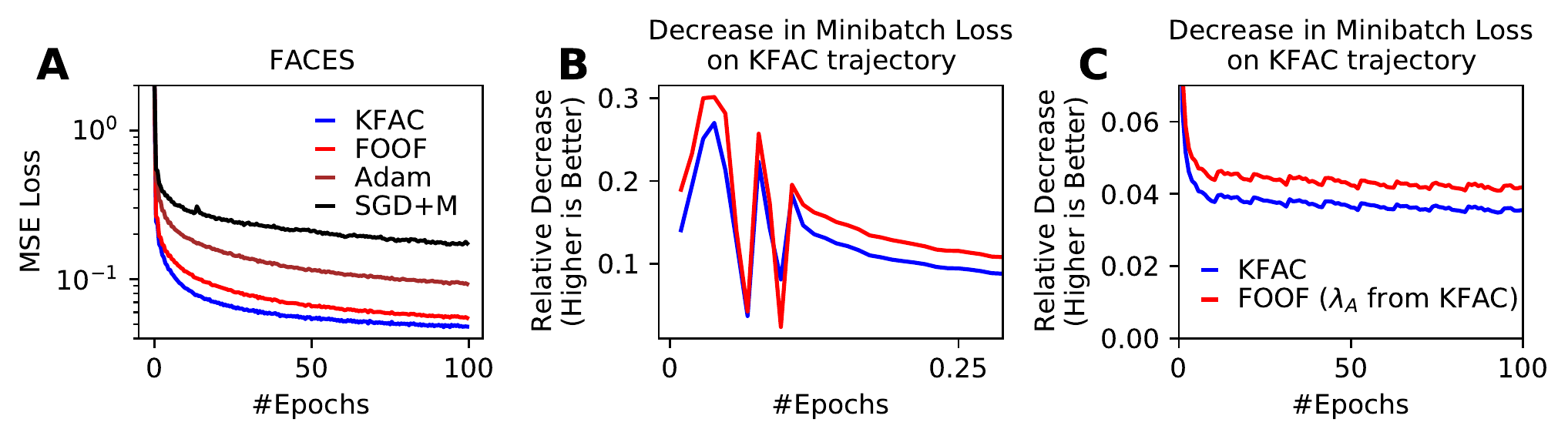}
	\vskip -15pt
	\caption{Performance for FACES autoencoder experiment. KFAC slightly outperforms FOOF, but when FOOF is on KFAC trajectory it typically makes more progress per udpate. This may suggest that the advantage of KFAC is due to choosing a different optimization trajectory. (B) shows same data as (C) with a different axes zoom.
	}
	\label{fig:faces}
\end{figure*}

\begin{figure*}[h]
	\centering
	\includegraphics[width=0.95\textwidth]{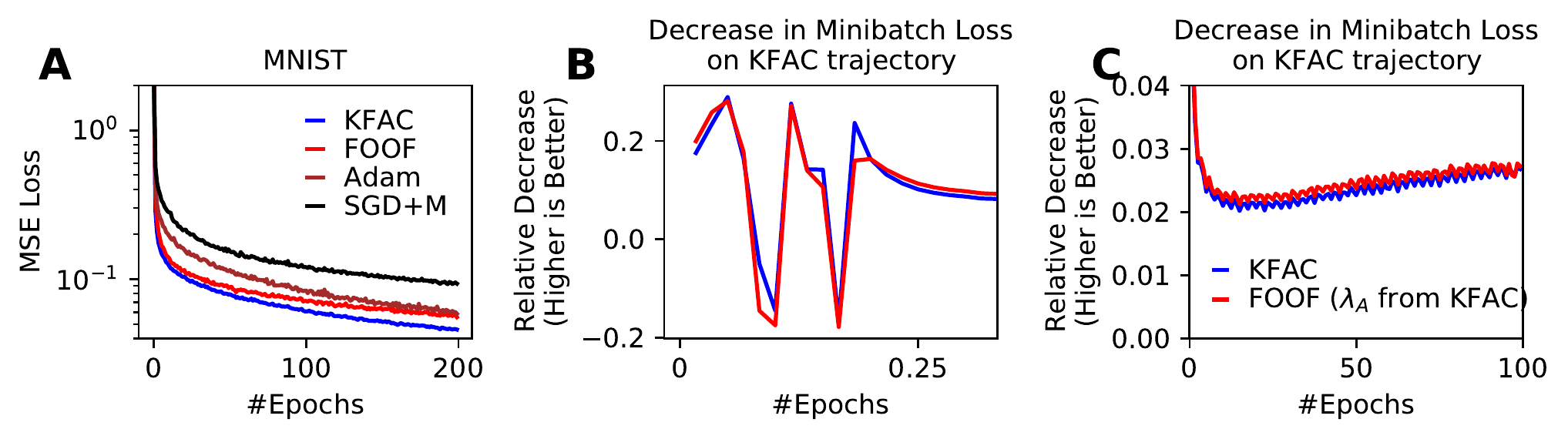}
	\vskip -15pt
	\caption{Performance for MNIST autoencoder experiment. KFAC slightly outperforms FOOF, but when FOOF is on KFAC trajectory it typically makes more progress per udpate. This may suggest that the advantage of KFAC is due to choosing a different optimization trajectory. (B) shows same data as (C) with a different axes zoom.
	}
	\label{fig:mnist}
\end{figure*}

\begin{figure*}[h]
	\centering
	\includegraphics[width=0.95\textwidth]{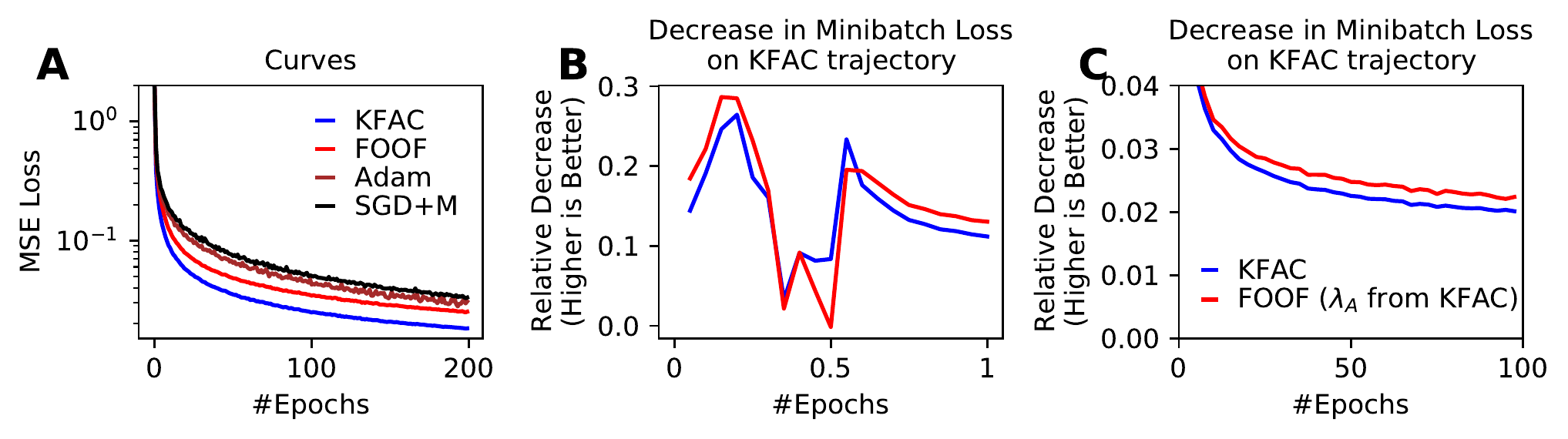}
	\vskip -15pt
	\caption{Performance for Curves autoencoder experiment. KFAC slightly outperforms FOOF, but when FOOF is on KFAC trajectory it typically makes more progress per udpate. This may suggest that the advantage of KFAC is due to choosing a different optimization trajectory. (B) shows same data as (C) with a different axes zoom.
	}
	\label{fig:curves}
\end{figure*}

\begin{figure*}[h]
	\centering
	\includegraphics[width=0.95\textwidth]{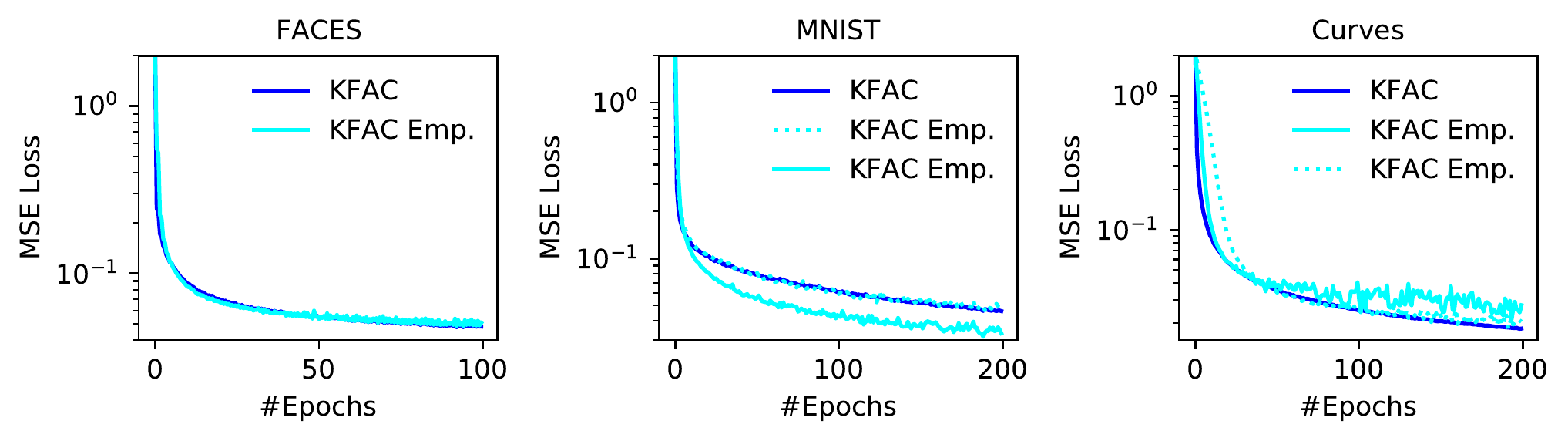}
	\vskip -15pt
	\caption{Performance of standard KFAC (using an MC sample to estimate the Fisher) and a version of KFAC using the empirical Fisher. Solid an dashed cyan lines show different hyperparametrisations of the same algorithm.
		The advantage fo the empirical Fisher on MNIST seems to be due to allowing different hyperparametrisations to be stable. 
	}
	\label{fig:auto_kfac_emp}
\end{figure*}

\clearpage
\section{Pseudocode, Implementation, Hyperparameters}
\label{sec:pseudo}
Pseudocode for FOOF is given in Algorithm~\ref{alg:foof}. Notation is anaologous to Section~\ref{sec:notation} and the amortisation described in Section~\ref{sec:exp_details}.

\textbf{Initialisation:} One detail omitted in the pseudocode is initialisation of $\Sig$ and $\mathbf{P}$. There is different ways to do this. We decided to perform Line~\ref{line:accumulate} of Algorithm~\ref{alg:foof} for a number of minibatches (50) before training and then executing line  Line~\ref{line:invert} once. In addition, we make sure the exponentially moving average is normalised.

\textbf{Amortisation Choices:} Amortising the overhead of FOOF is achieved by choosing $S, T$ suitably (large $T$ and small $S$ give the best runtimes). For fully connected layers updating $\Sig$ is cheap and we choose $S=T$ (i.e.\ $\Sig$ is updated at every step), we reported results for $T=1$ and $T=100$.
For the ResNet, computing $\A\A^T$ is more expensive and we chose $T=500$ (one inversion per epoch) and $S=10$, see also Appendix~\ref{sec:exp_details} as well as below.
Additional experiments (not shown) suggest that $\Sig$ can be estimated robustly on few datapoints and that it changes slowly during training.

\textbf{Hyperparameter Choices:} 
Note that a discussion of hyperparameter robustness is also provided in Section~\ref{sec:exp_details}.
We chose $m=0.95$ following \cite{martens2015optimizing}, brief experiments with $m=0.999$ seemed to give very similar results.  For damping $\lambda$ and learning rate $\eta$, we performed grid searches. It may be interesting that a bayesian interpretation of FOOF (details omitted) suggests choosing $\lambda$ as the precision used for standard weight initialisation schemes (e.g.\ Kaiming Normal initialisation) and seems to work well. If we choose this $\lambda$, we only need to tune the learning rate of FOOF, so that the required tuning is analogous to that of SGD. 
Alternatively, we typically fount $\lambda=100$ to work well, but the exact magnitude may depend on implementation details (in particular, how factors are scaled and how they depend on the batch size).

\textbf{Implementation and Convolutional Layers:} A PyTorch implementation of FOOF using for- and backward hooks is simple. The implementation, in particular computing $\A\A^T$, is most straightforward for fully connected layers, but can be extended to layers with parameter sharing. For example in CNNs, we can interpret the convolution as a standard matrix multiplication by ``extracting/unfolding'' individual patches (see e.g.\ \cite{grosse2016kronecker}) and then proceed as before. This is what our implementation does. 
A more efficient technique avoiding explicitly extracting patches (at the cost of making small approximations) is presented in \cite{ober2021global}.

\begin{algorithm}[tb]
	\caption{Gradient Descent on Neurons (FOOF)}
	\label{alg:foof}
	
	\begin{algorithmic}[1]
		\STATE {\bfseries Hyperparameters:} learning rate $\eta$, damping strength $\lambda$, exponential decay factor $m$, inversion period $T$, number of updates for input covariance $S\le T$ 
		\STATE {\bfseries Initialise:} $t=0$; For each layer $\ell$: Weights $\W_\ell$ (e.g. Kaiming-He init), exponential average $\Sig_\ell$ of $\A_\ell\A_\ell^T$ and its damped inverse $\mathbf{P}_\ell=(\Sig_\ell +\lambda\I)^{-1}$ (see Appendix~\ref{sec:pseudo} for details)
		\item[]
		\WHILE{train}
		\STATE Perform Standard Forward and Backward Pass For Current Mini-Batch With Loss $L$ 
		\FOR{each layer $\ell$}
		\STATE $\W_\ell \leftarrow \W_\ell - \eta \mathbf{P_\ell}\nabla_{\W_\ell} L$ \COMMENT{Update Parameters as in Eq. \ref{eq:foof_update}}
		\IF{$(t \text{ mod } T) == 0$} 
		\STATE $\mathbf{P}_\ell \leftarrow (\Sig_\ell + \lambda \I)^{-1}$ \COMMENT{Update Damped Inverse of Moving Average of $\A_\ell\A_\ell^T$ every $T$ steps} \label{line:invert}
		\ENDIF
		\IF{$((t +S) \text{ mod } T) \in \{0,\ldots, S-1\}$}
		\STATE $\Sig \leftarrow m\cdot \Sig_\ell + (1-m)\cdot\A_\ell\A_\ell^T$
		\COMMENT{Update Moving Average of $\A_\ell\A_\ell^T$ beginning $S$ steps \\ \hspace{138pt} before inversion in line~\ref{line:invert}. $\A_\ell$ is defined as in Section~\ref{sec:notation}.} \label{line:accumulate}
		\ENDIF
		\ENDFOR
		\STATE $t \leftarrow t+1$
		\ENDWHILE
	\end{algorithmic}
\end{algorithm}

\clearpage
\bibliographystyle{apalike}
\bibliography{sample}
\end{document}